\documentclass{article}
\usepackage{arxiv}

\usepackage[utf8]{inputenc} 
\usepackage[T1]{fontenc}    
\usepackage{hyperref}       
\usepackage{url}            
\usepackage{booktabs}       
\usepackage{amsfonts}       
\usepackage{nicefrac}       
\usepackage{microtype}      
\usepackage{lipsum}
\usepackage{graphicx}
\graphicspath{ {./images/} }

\usepackage{color,pifont}
\usepackage{caption}
\usepackage{subcaption}
\usepackage{multirow,arydshln}
\usepackage{nicematrix,enumitem} 
\usepackage{array}
\newcommand{\RN}[1]{%
  \textup{\uppercase\expandafter{\romannumeral#1}}%
}
\newcommand{\red}[1]{\textcolor{red}{#1}}
\newcommand{\blue}[1]{\textcolor{blue}{#1}}
\newcommand{\etal}{\textit{et al}.}
\newcommand{\wt}[1]{\textcolor{white}{#1}}

\title{Enhancing Image Rescaling using Dual Latent Variables in Invertible Neural Network}

\author{
 Min Zhang\thanks{Both authors contributed equally to this work when Min Zhang interned at Baidu.} \\
  University of Southern California\\ 
  Los Angeles, CA, USA\\ 
  \texttt{zhan980@usc.edu} \\
   \And
 Zhihong Pan$^*$ \\
  Baidu Research (USA) \\
  Sunnyvale, CA, USA \\
  \texttt{zhihongpan@baidu.com} \\
  \And
 Xin Zhou \\
  Baidu Research (USA)\\
  Sunnyvale, CA, USA \\
  \texttt{zhouxin16@baidu.com} \\
  \AND
 C.-C. Jay Kuo \\
  University of Southern California\\ 
  Los Angeles, CA, USA \\ 
  \texttt{cckuo@sipi.usc.edu}
}

\begin{document}
\maketitle
\begin{abstract}
Normalizing flow models have been used successfully for generative image super-resolution (SR) by approximating complex distribution of natural images to simple tractable distribution in latent space through Invertible Neural Networks (INN). These models can generate multiple realistic SR images from one low-resolution (LR) input using randomly sampled points in the latent space, simulating the ill-posed nature of image upscaling where multiple high-resolution (HR) images correspond to the same LR. Lately, the invertible process in INN has also been used successfully by bidirectional image rescaling models like IRN and HCFlow for joint optimization of downscaling and inverse upscaling, resulting in significant improvements in upscaled image quality. While they are optimized for image downscaling too, the ill-posed nature of image downscaling, where one HR image could be downsized to multiple LR images depending on different interpolation kernels and resampling methods, is not considered. A new downscaling latent variable, in addition to the original one representing uncertainties in image upscaling, is introduced to model variations in the image downscaling process.  This dual latent variable enhancement is applicable to different image rescaling models and it is shown in extensive experiments that it can improve image upscaling accuracy consistently without sacrificing image quality in downscaled LR images. It is also shown to be effective in enhancing other INN-based models for image restoration applications like image hiding.
\end{abstract}


\section{Introduction}
\begin{figure*}[t!]
 \begin{center}
     \includegraphics[width=\linewidth]{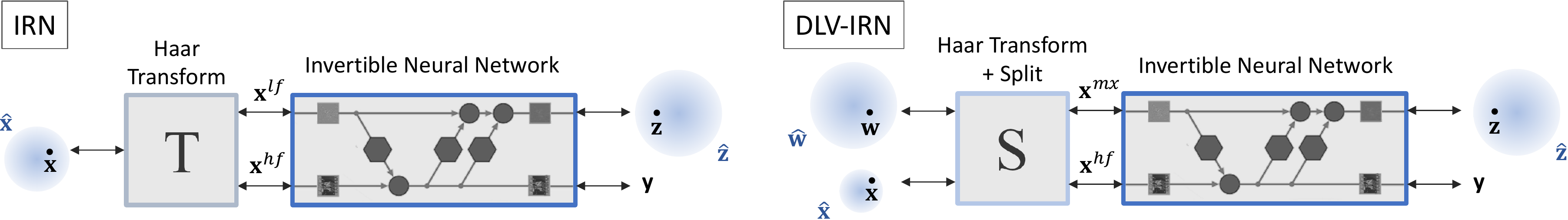}
 \end{center}

 \caption{Comparison between IRN \cite{xiao2020invertible} and the proposed DLV-IRN. $\mathbf{x}$ and $\mathbf{y}$ denote input HR image, output LR image, $\mathbf{w}$ and $\mathbf{z}$ denote the downscaling and upscaling latent variable respectively.  The corresponding $\hat{\mathbf{x}}$, $\hat{\mathbf{w}}$ and $\hat{\mathbf{z}}$ represent the variations caused by random sampling of $\mathbf{z}$ during inverse upscaling.  Note that superscripts $hr, lf$ refer to high and low frequency channels, and $mx$ is used for feature mixed from $\mathbf{x}$ and $\mathbf{w}$.}

 \label{fig:dlrn}
 \end{figure*}

Recent deep learning based image super-resolution (SR) methods have advanced the
performance of image upscaling significantly. These methods are only optimized for
the unidirectional upscaling process where the LR inputs are synthesized from a predefined downscaling kernel.
To take advantage of the potential mutual beneficiary reinforcement between downscaling and the inverse upscaling,
some image rescaling models \cite{kim_eccv_2018, sun_tip_2020, xiao2020invertible}
are developed to optimize the downscaling and the inverse upscaling processes jointly
and they yield significant improvements in accuracy of the upscaling task comparing to unidirectional SR models of the same
scale factors.  The state-of-the-art (SOTA) for learning based bidirectional image rescaling is set by the
invertible rescaling net (IRN) as proposed by Xiao \etal \cite{xiao2020invertible}.
As shown in Fig. \ref{fig:dlrn}, it is able to achieve the best performance so far as both the Haar transformation
and the invertible neural network (INN) \cite{ardizzone_iclr_2018} backbone are invertible processes which
fit naturally with the downscaling and inverse upscaling process.

Here the downscaling and inverse upscaling process of IRN can be described as
$\mathbf{y}, \mathbf{z} = f(\mathbf{x})$ and $\mathbf{x} = f^{-1}(\mathbf{y}, \mathbf{z})$ respectively.
When latent variable $\mathbf{z}$ is preserved, $\mathbf{x}$ can be perfectly restored as it is an invertible process.
This ideal inverse process does not exist in real applications when only LR output $\mathbf{y}$ is saved.
When the network is optimized to store as much information as allowed in $\mathbf{y}$
and transform $\mathbf{z}$ as an random variable independent of $\mathbf{y}$,
the restoration of  $\hat{\mathbf{x}}$ can be calculated as $f^{-1}(\mathbf{y}, \hat{\mathbf{z}})$ where $\hat{\mathbf{z}}$ is randomly sampled from a distribution like multivariate Gaussian,
illustrated as a colored area surrounding $\mathbf{z}$ in Fig. \ref{fig:dlrn}.  
After training, IRN is optimized to estimate the
same $\mathbf{x}$ from multiple $\hat{\mathbf{z}}$ samples.  For a given pair of $\mathbf{x}$ and $\mathbf{y}$, it is ideal to have the perfect restoration from multiple samples like
 \begin{equation}
 \mathbf{x} = f^{-1}(\mathbf{y}, \hat{\mathbf{z}}_i), i \in \mathbb{I},
 \label{eq:us1}
 \end{equation}
where $\mathbb{I}$ is a set with more than one element.

However, this is not feasible with IRN as it is self-conflicting with the invertible process.
Assume $|\,\mathbb{I}\,| > 1$, there must exist two different $\hat{\mathbf{z}}_i$ and $\hat{\mathbf{z}}_j$ that satisfy $f^{-1}(\mathbf{y}, \hat{\mathbf{z}}_i) = f^{-1}(\mathbf{y}, \hat{\mathbf{z}}_j) = \mathbf{x}$.
We can then see that $\mathbf{y}, \hat{\mathbf{z}}_i = \mathbf{y}, \hat{\mathbf{z}}_j$ after applying $f(\cdot)$ to both outputs.  Consequently,  $\hat{\mathbf{z}}_i$ and $\hat{\mathbf{z}}_j$ are identical and the size of set $\mathbb{I}$ must be 1.
In other words, for given $\mathbf{x}$ and $\mathbf{y}$, the average restoration error of $\hat{\mathbf{x}}$ from randomly sampled $\hat{\mathbf{z}}$ must be larger than zero as the zero-error prediction is only valid for one sample.
For reference, the distribution of $\hat{\mathbf{x}}$ is illustrated as an area surrounding $\mathbf{x}$ in Fig. \ref{fig:dlrn}.

 Aiming to resolve this self-conflicting problem, we introduce a new latent variable $\mathbf{w}$ during the forward downscaling process and
the enhanced invertible rescaling network with dual latent variables,  denoted as DLV-IRN, is also illustrated in Fig. \ref{fig:dlrn}
for comparison with IRN.  With the introduction of the second latent variable $\mathbf{w}$, the downscaling and upscaling
process are now denoted as $\mathbf{y}, \mathbf{z} = f(\mathbf{x}, \mathbf{w})$
and $\mathbf{x}, \mathbf{w} = f^{-1}(\mathbf{y}, \mathbf{z})$ respectively, 
assuming both images and latent variables are preserved.  In real applications where
the randomly sampled $\hat{\mathbf{z}}$ is used in the inverse upscaling process, the restoration of $\hat{\mathbf{x}}$ is described as
 \begin{equation}
 \begin{split}
 \hat{\mathbf{x}}, \hat{\mathbf{w}} & = f^{-1}(\mathbf{y}, \hat{\mathbf{z}}).
 \end{split}
 \label{eq:us2}
 \end{equation}
Similarly, the proposed DLV-IRN aims to restore the same $\mathbf{x}$ from multiple $\hat{\mathbf{z}}$ samples.
In other words, when $\mathbf{x}$ and $\mathbf{y}$ are given, it is ideal to have the exact same restoration from
different samples like
 \begin{equation}
 \mathbf{x}, \hat{\mathbf{w}}_j = f^{-1}(\mathbf{y}, \hat{\mathbf{z}}_j), j \in \mathbb{J},
 \label{eq:us3}
 \end{equation}
where $\mathbb{J}$ a set of ideal conditions.
With the introduction of $\mathbf{w}$, the size of set $\mathbb{J}$ is no longer limited to 1.
As explained later in Section \ref{sec:method}, assuming the invertible network is capable of unlimited learning,
it is theoretically possible to have the set of $\{\hat{\mathbf{z}}_j, j \in \mathbb{J}\}$ covers the full distribution
of $\hat{\mathbf{z}}$, resulting in perfect restoration of $\mathbf{x}$ in all conditions.
That is, the lower boundary of average restoration error between $\mathbf{x}$ and $\hat{\mathbf{z}}$ is zero, reduced from the original IRN.  

In practice, the proposed DLV-IRN is optimized using the following primary objective
 \begin{equation}
 f = \textstyle{\arg \min_{f_\theta}} \mathcal{L} (\mathbf{x}, f^{-1}_{\theta}(f_{\theta}(\mathbf{x}, \mathbf{w}), \hat{\mathbf{z}})),
 \label{eq:l2h}
 \end{equation}
where $\mathbf{w}$ and $\hat{\mathbf{z}}$ are randomly sampled for downscaling and upscaling respectively.
While the trained network is not capable of unlimited learning,
it is shown in comprehensive experiments the the dual latent variable enhancement can improve performances in bidirectional
image rescaling and other INN based image restoration models consistently.

In addition to reduction in lower boundary of the restoration error of $\hat{\mathbf{x}}$, the introduction of latent variable $\mathbf{w}$ enables the enhanced DLV-IRN
to model both aspects of the ill-posed nature of image rescaling.  First, the latent variable $\mathbf{z}$ included in previous works IRN \cite{xiao_eccv_2020}
and HCFlow \cite{liang2021hierarchical} is used to represent the high-frequency components which are needed in the upscaling scaling process
to restore the HR output $\mathbf{x}$ from the LR input $\mathbf{y}$.  For generative mode in HCFlow, the goal of random sampling in $\mathbf{z}$ is to
generate different $\mathbf{x}$, modeling the ill-posed nature of image upscaling where one LR input maps to multiple HR outputs.  For convenience,
here $\mathbf{z}$ is referred as the upscaling latent variable as it represents variations in high-frequency details which are needed for image upscaling.
On the other hand, to downscale an HR, we can generate multiple LR outputs
depending on different interpolation kernels and resampling methods, like nearest neighbour or bilinear interpolation.  The random sampling of $\mathbf{w}$
in training and testing of DLV-IRN simulate the ill-pose nature of image downscaling, where one HR input could map to different LR outputs.
Here $\mathbf{w}$ is referred as the downscaling latent variable as it represents variations in the image downscaling for a given HR input.

In summary, the main contributions of our work include:
\begin{itemize}
\setlength\itemsep{0.05em}
\item[$\bullet$] We are the first to propose including a downscaling latent variable $\mathbf{w}$ in invertible image rescaling models to improve baseline performance significantly without increased model complexity and sacrifice in LR image quality.

\item[$\bullet$] The dual latent variable scheme is also shown to be effective in enhancing other INN-based image restoration works like image steganography.

\end{itemize}

\section{Related Works}

\subsection{Invertible Neural Network}
The INN \cite{dinh2014nice,dinh2016density,ardizzone_iclr_2018,kingma2018glow,behrmann2019invertible} has an architecture of $f_\theta$ where its inverse function $f_\theta^{-1}$ share the same parameters,
leading to a cheaper inference.  Specifically, given an input $\mathbf{x}$, INN generates $z=f_\theta(\mathbf{x})$ in the forward pass, and $x$ can be recovered by $\mathbf{x}=f_\theta^{-1}(\mathbf{z})$.
In practice, for a complex distribution $p(\mathbf{x})$, $\mathbf{z}$ is commonly designed as an unobserved latent variable with a predefined tractable distribution.
As a result, the generative or reconstruction process in $\mathbf{x}$ can be modeled as random sampling in $\mathbf{x}$ and this model can be optimized using
standard SGD-based techniques as the negative log-likelihood (NLL) can be computed exactly.

INN was first proposed in NICE \cite{dinh2014nice}, a flow model by stacking non-linear additive coupling and other transformation layers. RealNVP \cite{dinh2016density} introduced multi-scale layers and substitute the non-linear additive coupling with affine coupling for lower computational cost and better regularization ability. Furthermore, the fixed permutation layer is replaced with 1 $\times$ 1 convolutional layers in Glow \cite{kingma2018glow}.  In contrast to previous unconditional generative models, lately INN have been applied to various conditional generative models like image SR \cite{lugmayr_eccv_2020}
and image colorization \cite{ardizzone_arxiv_2019}.

\subsection{Image Rescaling}
Image rescaling includes image downscaling and upscaling. Image downscaling resizes the HR image to a lower resolution which is visually pleasing. Frequency-based kernels \cite{mitchell1988reconstruction}, such as Bilinear and Bicubic, are commonly used for image downscaling. Image upscaling reconstructs promising HR image from the downscaled LR image, which is also known as image SR.  While powerful deep learning techniques have led to developments of
many image SR models \cite{dong_eccv_2014, kim_cvpr_2016_2, lim_cvprw_2017, zhang_cvpr_2018, zhang_eccv_2018} with impressive results, they rely on
LR inputs generated from predefined downscaling settings.

Recently, an encoder-decoder architecture \cite{kim_eccv_2018} is utilized as the first to jointly optimize the downscaling and upscaling process in bidirectional image rescaling.
Later a new content adaptive-resampler based image downscaling module \cite{sun_tip_2020} was proposed to train with existing differentiable SR models jointly.
However, these methods still suffer from the ill-posed problem, a visually plausible downscaled image may not be optimal for inverse upscaling. 
More recently, IRN \cite{xiao2020invertible} was proposed to use the bijective process in INN to
model downscaling and upscaling according to their reciprocal nature.
The proposed IRN is capable of generating a visually pleasing LR image and a latent variable $\mathbf{z}$
as well as restoring HR accurately from the saved LR and randomly sampled $\hat{\mathbf{z}}$.
While latent variable in IRN is regulated to be independent of the LR image, HCFlow \cite{liang2021hierarchical} proposed a hierarchical conditional framework so that
the high frequency components are conditional on the LR image hierarchically.
Later FGRN \cite{li2021approaching} proposed an encoder-decoder architecture to model downscaling and upscaling while using a separate invertible flow module, without using latent variables, as a guidance to learn the optimal
image downscaling in conjunction with image upscaling.

\subsection{Steganography and Image Hiding}

Steganography is the practice of hiding one message into a carrier, such as audio, image or video. Image hiding, specifically, is to unobtrusively place a whole image, i.e., secret image, within another image of the same size, i.e., cover image, and the secret image should be recovered from the stego image at the receiver end. 
Least Significant Bit (LSB) \cite{tamimi2013hiding} is a classic spatial domain method which uses the $n$ most significant bits of the secret image to replace the $n$ least significant bits of the cover image.  Newer methods use a similar way but hide information in frequency domains such as discrete wavelet transform (DWT) domain \cite{barni2001improved} so it is more undetectable.

Baluja \cite{baluja2017hiding} proposed the first deep learning solution for image hiding, which contains two sub-networks.
The concealing network hides the secret image into the cover image, outputting a stego image. Then the revealing network reconstructs the secret image from the stego image. The two sub-networks work as a pair and are trained simultaneously, but they do not share parameters so that the connection is loose, causing texture-copying artifacts and color distortion. 
Recently, HiNet \cite{jing2021hinet} utilized INN to solve the image hiding problem,  sharing the same set of network parameters for both image concealing and revealing, which improved the reconstruction accuracy significantly.
Besides, the secret information is hidden in the wavelet domain rather than pixel domain for high invisibility in the stego image.

\section{Proposed Method}
\label{sec:method}

\subsection{Preliminaries}
Flow-based models, which aim to learn a bijective mapping between the target space and the latent space, have been investigated for various applications.
For image generation models, the target space of HR images is modeled as a high-dimensional random variable $\mathbf{x}$ with a 
distribution of $\mathbf{x} \sim p(\mathbf{x})$.  The key aspect of flow models is the employment of
an invertible neural network (INN) $f_\theta$ that transforms $\mathbf{x}$ to a latent variable $\mathbf{z}$
with simple tractable distribution $\mathbf{z} \sim p(\mathbf{z})$ (e.g.\ Gaussian distribution).
Here $\theta$ represents the parameters of the invertible network which could be learned from training samples.
With the invertible network $f_\theta$, an HR image can be transformed to a latent variable as $\mathbf{z} = f_\theta(\mathbf{x})$,
and it can also be restored from a latent variable as $\mathbf{x} = f^{-1}_\theta(\mathbf{z})$.
Another key aspect of flow models is that, according to the change of variable formula,
the probability density $p(\mathbf{x} | \theta)$ can be calculated as
 \begin{equation}
 p(\mathbf{x} | \theta) = p(f_\theta(\mathbf{x})) \: \Big|\det \frac{\partial f_\theta}{\partial \mathbf{x}}(\mathbf{x})\Big|.
 \label{eq:px}
 \end{equation}
This allows the exact negative log-likelihood (NLL) $-\log p(\mathbf{x} | \theta)$ to be computed and
the network can then be trained by directly minimizing it using standard SGD-based techniques.

Lately, there have been investigations of conditional flow models, like using class [25] or image [3,49] as conditions
for image generation.  More recently, SRFlow
is proposed to generate realistic images from the condition of an LR image $\mathbf{y}$.
Similar to Equation \ref{eq:px}, the conditional probability density of $\mathbf{x}$ is calculated as
 \begin{equation}
 p(\mathbf{x} | \mathbf{y},\theta) = p(f_\theta(\mathbf{x}; \mathbf{y})) \: \Big|\det \frac{\partial f_\theta}{\partial \mathbf{x}}(\mathbf{x}; \mathbf{y})\Big|.
 \label{eq:pxy}
 \end{equation}
Using a large set of HR-LR training pairs $\{(\mathbf{x}_i, \mathbf{y}_i)\}^M_{i=1}$,
the invertible network can be trained by minimizing the NLL loss using a data-driven process.
After training, the conditional distribution $p(\mathbf{x} | \mathbf{y}, \theta)$ captures the nature of all possible realistic HR images corresponding to a known $\mathbf{y}$
and multiple SR images can be generated from one LR reference $\mathbf{y}$ and randomly sampled latent variable $\mathbf{z}$.

Most recently, HCFlow aims to use the bijective transformation of flow models to
model two different modes: SR generation from LR and bidirectional image rescaling.
In the latter case, $\mathbf{y}$ is not the known LR ground-truth (GT) used as a condition, but part of outputs in the latent space
as $\mathbf{x} \leftrightarrow (\mathbf{y}, \mathbf{a}) = f_\theta(\mathbf{x})$ where $\mathbf{a}$ is an intermediate variable representing
decomposed high-frequency components from $\mathbf{x}$.
To explain the role of the flow model in transforming the complex distribution
of natural images $\mathbf{x}$ to tractable distribution in latent space, a GT LR $\overline{\mathbf{y}}$ is introduced as a condition as below
 \begin{equation}
 p(\mathbf{x}|\overline{\mathbf{y}}) \leftrightarrow p(\mathbf{y},\mathbf{a}|\overline{\mathbf{y}}) = p(\mathbf{a}|\mathbf{y},\overline{\mathbf{y}})p(\mathbf{y}|\overline{\mathbf{y}}).
 \label{eq:pxyb}
 \end{equation}
Ideally, the model is trained to generate the LR image $\mathbf{y}$ exactly as the GT LR $\overline{\mathbf{y}}$,
so the first factor of the above calculation is simplified as $p(\mathbf{a}|\mathbf{y})$,
which is further mapped to a standard multivariate Gaussian distribution in latent space: $p(\mathbf{z}) = \mathcal{N}(\mathbf{z}|\mathbf{0}, \mathbf{I})$.
And the second factor
could be also formulated as $\delta(\mathbf{y}$ - $\overline{\mathbf{y}})$, which can be further
approximated by a multivariate Gaussian distribution as $\lim_{\Sigma \rightarrow 0} \mathcal{N}(\mathbf{x}|\overline{\mathbf{y}}, \Sigma)$.
As a result, the complex distribution $p(\mathbf{x}|\overline{\mathbf{y}})$ becomes the product of two tractable Gaussian distributions.
Note that $\mathbf{a}$ is introduced to make the transformation from and to latent variable $\mathbf{z}$ conditional to $\overline{\mathbf{y}}$
for generative SR models as they needed to be trained by maximum likelihood estimation (MLE).
 
\subsection{Dual Latent Variables}
\label{sec:dal}
While the newly introduced intermediate variable $\mathbf{a}$ in HCFlow makes it possible
to share the same invertible network architecture for both generative image SR and bidirectional image rescaling,
it is not really needed for image rescaling as the NLL loss is not used in actual training. 
Alternatively, IRN works on the bidirectional image rescaling problem exclusively.  It uses the INN backbone to transform
the input image $\mathbf{x}$ to an LR output $\mathbf{y}$ and latent variable $\mathbf{z}$
without intermediate $\mathbf{a}$.  By training the network to make $\mathbf{z}$
a Gaussian distribution independent of $\mathbf{y}$, an accurate restoration of $\mathbf{x}$ is made possible using randomly sampled
$\hat{\mathbf{z}}$ during the inverse upscaling process.
Similar to Equation \ref{eq:pxyb}, for IRN, the distribution transformation process becomes $p(\mathbf{x}) \leftrightarrow p(\mathbf{y})p(\mathbf{z})$.

While both SRFlow and IRN use similar INN backbones to model bijective transformation between the image space and the latent space and achieves great performance
in generative image SR and image rescaling respectively, there is a distinctive difference in the purpose of random sampling in the latent space.
For SRFlow, the random sampling of $\mathbf{z}$ is beneficial to generate diverse versions of SR output from one LR image.
This is also aligned with one aspect of the ill-posed nature in image SR application, as there are multiple HR images corresponding to one LR image.
In other words, $\mathbf{z}$ is a latent variable for image upscaling, which represents high-frequency components of HR images that can only
be observed after the image upscaling process of $f^{-1}_\theta(\cdot)$ is applied.
While for IRN in ideal situations, as its goal is to restore HR image exactly,
the random sampling of $\mathbf{z}$ works against the goal of restoring the same HR result $\mathbf{x}$ when $\mathbf{y}$ is known.
When $\mathbf{x}$ is given, $\mathbf{y}$ and $\mathbf{z}$ are also determined as $(\mathbf{y}, \mathbf{z}) = f_\theta(\mathbf{x})$.
However, since $\mathbf{z}$ is invisible for the inverse upscaling process, a randomly sampled $\hat{\mathbf{z}}$
is used instead.
As explained in the introduction, different sampling of $\hat{\mathbf{z}}$ must map with different $\hat{\mathbf{x}}$ due to the invertible nature of the network.
That is, if $\hat{\mathbf{z}}_i = \mathbf{z}$, then for any $j \neq i$, we can see that $\hat{\mathbf{x}_j} \neq \hat{\mathbf{x}_i} = \hat{\mathbf{x}}$
so $\|\mathbf{x}-\hat{\mathbf{x}}_j\|_2 > 0$.
For a given $\mathbf{x}$ and any $N > 1$, the expected restoration error of $N$ randomly sampled $\hat{\mathbf{x}}$ has a positive lower boundary as
 \begin{equation}
 \mathbb{E}\big[\|\mathbf{x}-\hat{\mathbf{x}}\|_2\big] > 0.
 \label{eq:err1}
 \end{equation}

To alleviate this self-conflicting nature between random sampling of $\hat{\mathbf{z}}$ and exact restoration of $\mathbf{x}$,
here we propose a second latent variable $\mathbf{w}$ as additional input
besides $\mathbf{x}$.  The forward downscaling and backward upscaling process become
 \begin{equation}
 \begin{split}
 \mathbf{y}, \mathbf{z} & = f_\theta(\mathbf{x}, \mathbf{w}), \\
 \mathbf{x}, \hat{\mathbf{w}} & = f^{-1}_\theta(\mathbf{y}, \hat{\mathbf{z}}),
 \end{split}
 \label{eq:w}
 \end{equation}
where $\mathbf{w}$ is randomly sampled from an independent normal distribution for downscaling while $\hat{\mathbf{z}}$ is randomly sampled for subsequent upscaling.
Using the same optimization objective of minimizing $\mathbb{E}\big[\|\mathbf{x}-\hat{\mathbf{x}}\|_2\big]$ for this random
downscaling and upscaling process, we can show that, assuming the learning capability of $f_\theta(\cdot)$ is not limited,
its lower boundary is zero without conflicting with the invertible network characteristics.
For any given $\mathbf{x}$, we can generate $\mathbf{y}$ as its corresponding LR output.
For any set of $N$ unique values of $\mathbf{z}$: $\mathbb{Z} = \{\mathbf{z}_i | i=1, 2, \dots, N\}$ and another set of $N$ unique elements $\mathbb{W} = \{\mathbf{w}_i | i=1, 2, \dots, N\}$, there
exists a invertible network $f_{\theta_N}(\cdot)$ that satisfies
 \begin{equation}
 \begin{split}
 \mathbf{y}, \mathbf{z}_i & = f_{\theta_N}(\mathbf{x}, \mathbf{w}_i), \\
 \mathbf{x}, \mathbf{w}_i & = f^{-1}_{\theta_N}(\mathbf{y}, \mathbf{z}_i)
 \end{split}
 \label{eq:wz}
 \end{equation}
for any $i \in [1..N]$.  Due to the unlimited learning ability of $f_\theta(\cdot)$, we can see that
 \begin{equation}
 \lim_{N\to\infty} \mathbb{E}\big[\|\mathbf{x}-\hat{\mathbf{x}}\|_2\big] = 0.
 \label{eq:lim}
 \end{equation}

In summary, with the introduction of $\mathbf{w}$, the lower boundary of the restoration error is reduced from a positive value to zero theoretically for any big $N$.
While the learning ability of the invertible network is limited in practice, it is shown in extensive experiments that it helps
reducing the restoration error for different applications.
Alternatively, the introduction of $\mathbf{w}$ allows the modeling of the downscaling aspect of the ill-posed nature of image rescaling.  That is,
due to variations in blur kernels and resampling methods, there exist various LR outputs from a single HR input.  With two latent variables,
$\mathbf{w}$ for image downscaling and $\mathbf{z}$ for upscaling, our proposed method can be applied to different INN based image rescaling models
to become an enhanced dual latent variable (DLV) version for improved performance.


\begin{figure*}[t!]
\centering
\includegraphics[width=1\linewidth]{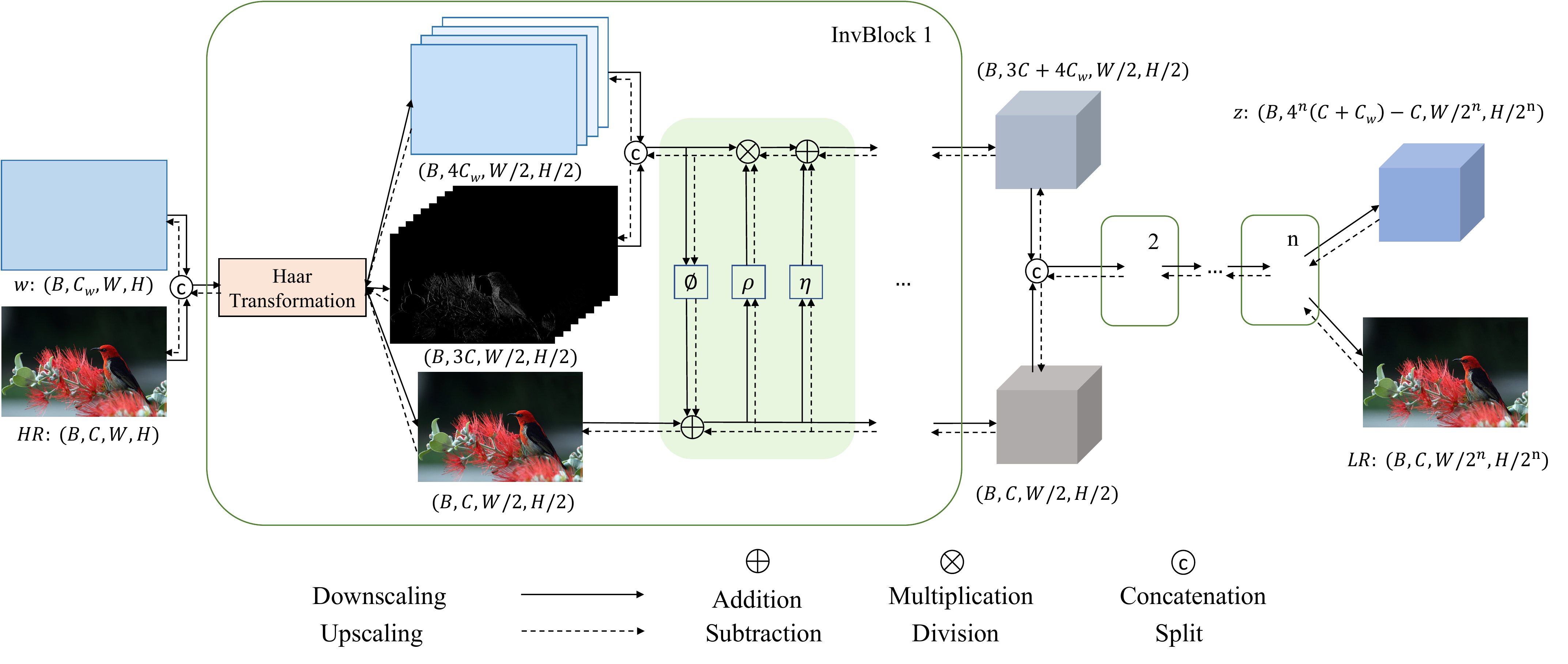}
\caption{Overview of the dual latent variables in the IRN method. $n=1$ and $n=2$ for a downscaling factor of 2 and 4, respectively. Haar Transformation divides the input into low-frequency and high-frequency components.}
\label{fig:pipeline}
\end{figure*}

\subsection{Model Architecture}
The introduction of the downscaling latent variable $\mathbf{w}$ only causes minor changes in model architecture of the baseline model.
Using IRN as the primary baseline model here the model architecture of the enhanced DLV-IRN is illustrated in Fig. \ref{fig:pipeline}.  In addition to the original HR input $\mathbf{x}$ , the new latent variable $\mathbf{w}$ is introduced for each pixel. 
After both are applied with Haar transformation, the low-frequency components of $\mathbf{x}$ are split and preserved as the low-frequency branch while the remaining high-frequency components are concatenated with all channels from $\mathbf{w}$ as the mixture branch.
Other than increased channel numbers, the transformation networks between the low-frequency and the mixture branches are kept the same as the baseline model.
Depending on the scaling factor, there could be more than one InvBlock cascaded to build the full pipeline, with the split and concatenating step applied once at the beginning of each InvBlock.   At the end, the output of the low-frequency branch is $\mathbf{y}$ and the one from the mixture branch is $\mathbf{z}$.
The image and feature dimensionality at different stages are included in Fig. \ref{fig:pipeline} for reference.

\subsection{Training Objectives}

To better demonstrate the effectiveness of the innovative dual latent variable module itself, it is desired to keep the training objective of original baseline model unchanged for fair comparison. 
In the case of newly introduced latent variable $\mathbf{w}$, as it is randomly sampled during
the forward downscaling process and could be safely discarded after upscaling, it is possible to use the same overall loss as IRN for training DLV-IRN, denoted as
\begin{equation}
     L = \lambda_1 L_{r}+\lambda_2 L_{g}+\lambda_3 L_{d}+\lambda_4 L_{i}.
\label{eq:loss}
\end{equation}
Here $L_{r}$ is the $L1$ reconstruction loss for upscaled HR output $\hat{\mathbf{x}}$ and
$L_{g}$ is the $L2$ guidance loss for downscaled LR output $\mathbf{y}$ in reference to
a downsampled LR reference $\overline{\mathbf{y}}$ using bicubic interpolation.
For $L_{d}$, same as in IRN \cite{xiao_eccv_2020}, the partial distribution  matching loss $-\log p(\mathbf{z})$ is
used for stable training.  These three are similar to the baseline IRN and their weights are also kept unchanged for consistency.
The last term $L_{i}$ is an LR invariance loss introduced for the newly proposed downscaling later variable $\mathbf{W}$.
As the reverse upscaling output $\hat{\mathbf{w}}$ has no impact on model performance, the LR invariance loss is introduced to the forward downscaling process only.
When $\mathbf{x}$ is given and $\mathbf{w}_j, j \in [1, m]$ is randomly sampled, $L_i$ is designed to make the output LR $\mathbf{y}_j$ invariant to $\mathbf{w}_j$ and it is calculated
as
\begin{equation}
     L_{i} = \sqrt{\frac{1}{m-1} \sum_{j=1}^{m} (\mathbf{y}_j - \tilde{\mathbf{y}})^2}
\label{eq:li}
\end{equation}
where $\tilde{\mathbf{y}}$ is the average LR output. 
This loss does not rely on any supervision from LR references and works together with $L_g$ as an enhanced semi-supervised learning in LR output.  In our experiments, $m$ is set as 3 and $\lambda_4 = s^2/4$ where $s$ is the scaling factor.

\section{Experiments}

\subsection{Experimental Setup}

\noindent\textbf{Dataset and metrics.}
Following the IRN baseline, we train our DLV-IRN model on the DIV2K \cite{agustsson_ntire_2017} dataset, which contains 800 2K resolution training images.
For the DLV-HCFlow, a combined DF2K dataset including both DIV2K and Flickr2K \cite{timofte_ntire_2017} is also used to compare with the HCFlow baseline.
For image rescaling, we evaluate our method on five standard benchmark datasets, i.e., the Set5 \cite{bevilacqua_bmvc_2012}, Set14 \cite{zeyde_iccs_2010}, BSD100 \cite{martin_iccv_2001}, Urban100 \cite{huang_cvpr_2015} and the validation set of DIV2K. For image hiding, the testing datasets include the validation set of DIV2K with 100 1024 $\times$ 1024 images, ImageNet \cite{russakovsky2015imagenet} with 50,000 256 $\times$ 256 images, and the COCO \cite{lin2014microsoft} dataset with 5,000 256 $\times$ 256 images. Following IRN \cite{xiao2020invertible}, PSNR and SSIM \cite{wang_tip_2004} on the Y channel of the YCbCr color space are used for assessing upscaled image quality. Since the downscaled LR images do not have ground-truth, we employ NIQE \cite{mittal_spl_2012} and PIQE \cite{venkatanath_nccc_2015} which are non-reference metrics in addition to SSIM. For image hiding, we also compare MAE and RMSE by following HiNet \cite{jing2021hinet}. The larger value of PSNR, SSIM and the smaller value of NIQE, PIQE, MAE and RMSE represent better image quality.


\noindent\textbf{Image rescaling.}
Two baseline models, IRN and HCFlow, are used to assess the effectiveness of dual latent variable enhancement.  For both DLV-IRN and DLV-HCFlow, the downscaling latent variable $\mathbf{w}$ is added as
a 2-channel pixel-wise variable.  The model size change introduced by this channel addition is insignificant.  To verify that the performance improvements brought by DLV enhancement is not caused by simple
model size change, IRN and HCFlow are also augmented in model depth, denoted as IRN$^\dagger$ and HCFlow$^\dagger$, to match their DLV counterparts in parameter numbers for fair comparison.
For both the augmented-depth and DLV variants, they are trained using the same settings as their corresponding baselines, including loss weights, learning rate and number of iterations.

\noindent\textbf{Image hiding.}
We follow most of the experimental settings in HiNet \cite{jing2021hinet} except for the following aspects. First, observing that the original HiNet has gradient explosion during training, we change the learning rate to be $2 \times 10^{-4}$, halved every 1,000 epochs, and add gradient clipping. Second, the lack of quantization in HiNet causes unreliable results so we add a quantization step before the revealing process. Third, with the above changes, we find that the model can learn more after 80K iterations so the total number of epochs is set to be 5,000. For DLV-HiNet, the 2-channel pixel-wise downscaling latent variable $\mathbf{w}$ is split to the cover and secret branch separately, each with 1-channel. For fair comparison, we retrained the model of Baluja \cite{baluja2017hiding} based on a third party implementation~\footnote{\url{https://github.com/arnoweng/PyTorch-Deep-Image-Steganography}} using the same settings mentioned above for consistency.

\begin{table}[t!]
    \centering
    \caption{Comparison of $\times 4$ HR image reconstruction results (PSNR) for different model hyperparameter and settings. The best two results highlighted in \red{red} and \blue{blue} respectively.}
    \label{tab:ablation}
    \vspace{10pt}
    \begin{tabular}{c|cc|c|c|c|c|c|c} \hline
        $C_\mathbf{w}$ & $\hat{\mathbf{z}}$ & $\mathbf{w}$ & $L_i$ & Set5 & Set14 & BSD100 & Urban100 & DIV2K \\ \hline\hline
        \multirow{3}*{2} & $\hat{\mathbf{z}}_\mathcal{N}$ & $\mathbf{w}_\mathcal{N}$ & \multirow{6}*{$\times$} & 36.02 & 32.55 & 31.49 & 31.07 & 34.88 \\
        & $\hat{\mathbf{z}}_\mathcal{N}$ & $\mathbf{w}_0$ & & 36.02 & 32.48 & 31.43 & 31.05 & 34.82 \\ 
        & $\hat{\mathbf{z}}_0$ & $\mathbf{w}_0$ & & 36.33 & 32.90 & 31.73 & 31.68 & 35.19 \\  \cdashline{1-3} \cline{5-9}
        1 & \multirow{4}*{$\hat{\mathbf{z}}_0$} & \multirow{4}*{$\mathbf{w}_\mathcal{N}$} & & 36.29 & 32.83 & 31.74 & 31.72 & 35.21 \\ 
        2 &  &  & & \red{36.43} & \blue{32.98} & \blue{31.82} & \blue{31.92} & \blue{35.29} \\
        3 &  &  & & 36.38 & 32.97 & 31.78 & 31.78 & 35.25 \\ \cdashline{4-4} \cline{5-9}
        2 &  &  & \checkmark & \blue{36.42} & \red{33.03} & \red{31.84} & \red{32.06} & \red{35.34} \\ \hline
    \end{tabular}
\end{table}


\subsection{Ablation Study}
For the image rescaling models studied here, while the HR image can consistently restored from random samples of $\hat{\mathbf{z}}$, we only need one specific $\hat{\mathbf{z}}_k$ to restoration during testing.
In fact, while the random sampling of $\hat{\mathbf{z}}$ is critical for diversity in generative models, it is not beneficial to have uncertainty in restored HR image.
As the distribution matching loss $L_d$ is applied to $\mathbf{z}$ and there is no direct constraint on selection of $\hat{\mathbf{z}}$, one practical choice is to keep it consistent across training and testing.
As validated in FGRN \cite{li2021approaching}, for IRN, fixing $\hat{\mathbf{z}}$ as a constant zero $\hat{\mathbf{z}}_0$ for both training and testing could achieve equivalent performance.
In the case of our DLV-IRN, using constant $\hat{\mathbf{z}}_0$ is also beneficiary for stable training as random sampling of both $\hat{\mathbf{z}}$ and $\mathbf{w}$ could cause oscillation in joint optimization.
For comparison, an ablation study is conducted to compare the effects of different settings in both $\hat{\mathbf{z}}$ and $\mathbf{w}$, where $\hat{\mathbf{z}}_0$ and $\mathbf{w}_0$ refer to the constant zero
and $\hat{\mathbf{z}}_\mathcal{N}$ and $\mathbf{w}_\mathcal{N}$ represent random sampling from a normal distribution.  All model variants here are trained for $250K$ iterations with an initial learning
rate of $2 \times 10^{-4}$, halved after every $50K$ iterations.  As shown in Table \ref{tab:ablation}, using constant $\hat{\mathbf{z}}_0$ is clearly better than $\hat{\mathbf{z}}_\mathcal{N}$, while
random sampling of $\mathbf{w}_\mathcal{N}$ is slightly better than $\mathbf{w}_0$.

Additionally, for different latent variable channel number $C_\mathbf{w}$,
it is shown that there is consistent performance gain when $C_\mathbf{w}$
increases from 1 to 2, but the overall performance drops when it is further increased to 3.
Lastly, the addition of $L_i$ loss, which is only implemented for DLV-IRN, is shown to further improve HR reconstruction performance.
Due to limited space, results with $L_i$ included in Table \ref{tab:ablation} are for the default setting of $\mathbf{w}_\mathcal{N}$,
$\hat{\mathbf{z}}_0$ and $C_\mathbf{w}=2$ only.

\subsection{Experiments on Image Rescaling}

\begin{table*}[t!]
    \centering
    \caption{Quantitative results of upscaled HR image quality from different rescaling methods on benchmark datasets.}
    \label{tab:hr}
    \resizebox{\textwidth}{!}{%
    \begin{NiceTabular}{c|c|c|c|cc|cc|cc|cc|cc} \hline
         \multirow{2}*{Type} & Downscaling \& & \multirow{2}*{Scale} & \multirow{2}*{Param} & \multicolumn{2}{c|}{Set5} & \multicolumn{2}{c|}{Set14} & \multicolumn{2}{c|}{BSD100} & \multicolumn{2}{c|}{Urban100} & \multicolumn{2}{c}{DIV2K} \\ \cline{5-14}
         & Upscaling & & & PSNR$\uparrow$ & SSIM$\uparrow$ & PSNR$\uparrow$ & SSIM$\uparrow$ & PSNR$\uparrow$ & SSIM$\uparrow$ & PSNR$\uparrow$ & SSIM$\uparrow$ & PSNR$\uparrow$ & SSIM$\uparrow$ \\ \hline\hline
         \multirow{4}*{\RN{1}} & Bicubic \& Bicubic & \multirow{4}*{$\times$ 2} & - & 33.66 & 0.9299 & 30.24 & 0.8688 & 29.56 & 0.8431 & 26.88 & 0.8403 & 31.01 & 0.9393 \\
         & Bicubic \& EDSR \cite{lim_cvprw_2017} & & 40.7M & 38.20 & 0.9606 & 34.02 & 0.9204 & 32.37 & 0.9018 & 33.10 & 0.9363 & 35.12 & 0.9699 \\
         & Bicubic \& RCAN \cite{zhang_eccv_2018} & & 15.4M & 38.27 & 0.9614 & 34.12 & 0.9216 & 32.41 & 0.9027 & 33.34 & 0.9384 & 35.04 & 0.9405 \\
         & Bicubic \& SAN \cite{dai_cvpr_2019} & & 15.7M & 38.31 & 0.9620 & 34.07 & 0.9213 & 32.42 & 0.9028 & 33.10 & 0.9370 & 36.73 & 0.9497 \\ \hline
         \multirow{5}*{\RN{2}} & CAR \& EDSR \cite{sun_tip_2020} & \multirow{5}*{$\times$ 2} & 51.1M & 38.94 & 0.9658 & 35.61 & 0.9404 & 33.83 & 0.9262 & 35.24 & 0.9572 & 38.26 & 0.9599 \\
         & IRN \cite{xiao_eccv_2020} & & 1.66M & 43.99 & 0.9871 & 40.79 & 0.9778 & 41.32 & 0.9876 & 39.92 & 0.9865 & 44.32 & 0.9908 \\
         & FGRN \cite{li2021approaching} & & 1.33M & \blue{44.15} & \blue{0.9902} & \red{42.28} & \red{0.9840} & \blue{41.87} & \blue{0.9887} & \red{41.71} & \blue{0.9904} & \blue{45.08} & \blue{0.9917} \\
         & \tabularnote{IRN$^\dagger$ is a variant of IRN with increased model depth for fair comparison with DLV-IRN in terms of number of parameters.}IRN$^\dagger$ & & 2.08M & 43.91 & 0.9871 & 40.34 & 0.9777 & 41.25 & 0.9875 & 39.80 & 0.9863 & 44.23 & 0.9908\\ 
         & DLV-IRN (ours) & & 1.89M & \red{45.42} & \red{0.9910} & \blue{42.16} & \blue{0.9839} & \red{42.91} & \red{0.9916} & \blue{41.29} & \red{0.9904} & \red{45.58} & \red{0.9934} \\ \hline\hline
         \multirow{4}*{\RN{1}} & Bicubic \& Bicubic & \multirow{4}*{$\times$ 4} & - & 28.42 & 0.8104 & 26.00 & 0.7027 & 25.96 & 0.6675 & 23.14 & 0.6577 & 26.66 & 0.8521 \\
         & Bicubic \& EDSR \cite{lim_cvprw_2017} & & 43.1M & 32.62 & 0.8984 & 28.94 & 0.7901 & 27.79 & 0.7437 & 26.86 & 0.8080 & 29.38 & 0.9032 \\
         & Bicubic \& RCAN \cite{zhang_eccv_2018} & & 15.6M & 32.63 & 0.9002 & 28.87 & 0.7889 & 27.77 & 0.7436 & 26.82 & 0.8087 & 30.77 & 0.8460 \\
         & Bicubic \& SAN \cite{dai_cvpr_2019} & & 15.7M & 32.64 & 0.9003 & 28.92 & 0.7888 & 27.78 & 0.7436 & 26.79 & 0.8068 & 31.14 & 0.8510 \\ \hline
         \multirow{6}*{\RN{2}} & CAR \& EDSR \cite{sun_tip_2020} & \multirow{6}*{$\times$ 4} & 52.8M & 33.88 & 0.9174 & 30.31 & 0.8382 & 29.15 & 0.8001 & 29.28 & 0.8711 & 32.82 & 0.8837 \\
         & IRN \cite{xiao_eccv_2020} & & 4.35M & 36.19 & 0.9451 & 32.67 & 0.9015 & 31.64 & 0.8826 & 31.41 & 0.9157 & 35.07 & 0.9318 \\
         & HCFlow \cite{liang2021hierarchical} & & 4.40M & 36.29 & 0.9468 & 33.02 & 0.9065 & 31.74 & 0.8864 & 31.62 & 0.9206 & \blue{35.23} & \blue{0.9346} \\
         & FGRN \cite{li2021approaching} & & 3.35M & \red{36.97} & \red{0.9505} & \red{33.77} & \red{0.9168} & \blue{31.83} & \red{0.8907} & \blue{31.91} & \red{0.9253} & 35.15 & 0.9322 \\
         & IRN$^\dagger$ & & 5.44M & 36.20 & 0.9445 & 32.33 & 0.8986 & 31.64 & 0.8808 & 31.51 & 0.9152 & 35.07 & 0.9308 \\ 
         & DLV-IRN (ours) & & 5.49M & \blue{36.62} & \blue{0.9484} & \blue{33.26} & \blue{0.9093} & \red{32.05} & \blue{0.8893} & \red{32.26} & \blue{0.9253} & \red{35.55} & \red{0.9363} \\ \hline
    \end{NiceTabular}}
\end{table*}

\begin{figure*}[t!]
\captionsetup[subfigure]{font=footnotesize, labelformat=empty}
\begin{center}
\begin{subfigure}[b]{0.15\textwidth}
    \centering
    \includegraphics[width=\textwidth, interpolate=false]{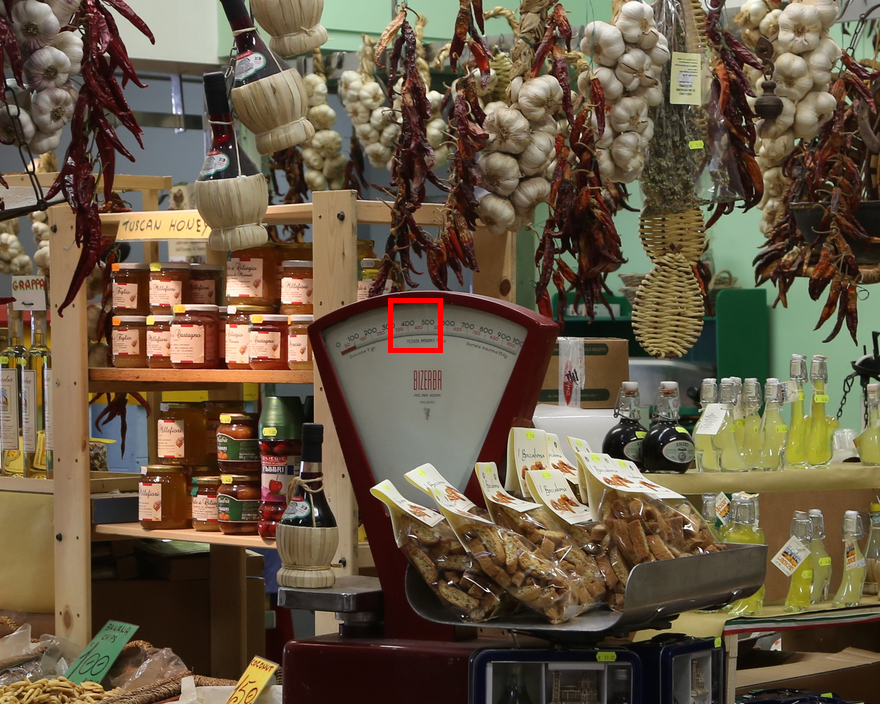}
\end{subfigure} \hspace*{-0.4em}
\begin{subfigure}[b]{0.12\textwidth}
    \centering
    \includegraphics[width=\textwidth, interpolate=false]{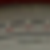}
\end{subfigure} \hspace*{-0.4em}
\begin{subfigure}[b]{0.12\textwidth}
    \centering
    \includegraphics[width=\textwidth, interpolate=false]{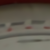}
\end{subfigure} \hspace*{-0.4em}
\begin{subfigure}[b]{0.12\textwidth}
    \centering
    \includegraphics[width=\textwidth, interpolate=false]{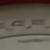}
\end{subfigure} \hspace*{-0.4em}
\begin{subfigure}[b]{0.12\textwidth}
    \centering
    \includegraphics[height=\textwidth, interpolate=false]{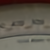}
    \end{subfigure} \hspace*{-0.4em}
\begin{subfigure}[b]{0.12\textwidth}
    \centering
    \includegraphics[width=\textwidth, interpolate=false]{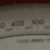}
\end{subfigure} \hspace*{-0.4em}
\begin{subfigure}[b]{0.12\textwidth}
    \centering
    \includegraphics[width=\textwidth, interpolate=false]{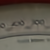}
\end{subfigure} \hspace*{-0.4em}
\begin{subfigure}[b]{0.12\textwidth}
    \centering
    \includegraphics[width=\textwidth, interpolate=false]{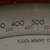}
\end{subfigure} \hspace*{-0.4em}

\begin{subfigure}[b]{0.15\textwidth}
    \centering
    \includegraphics[width=\textwidth, height=0.8\textwidth, interpolate=false]{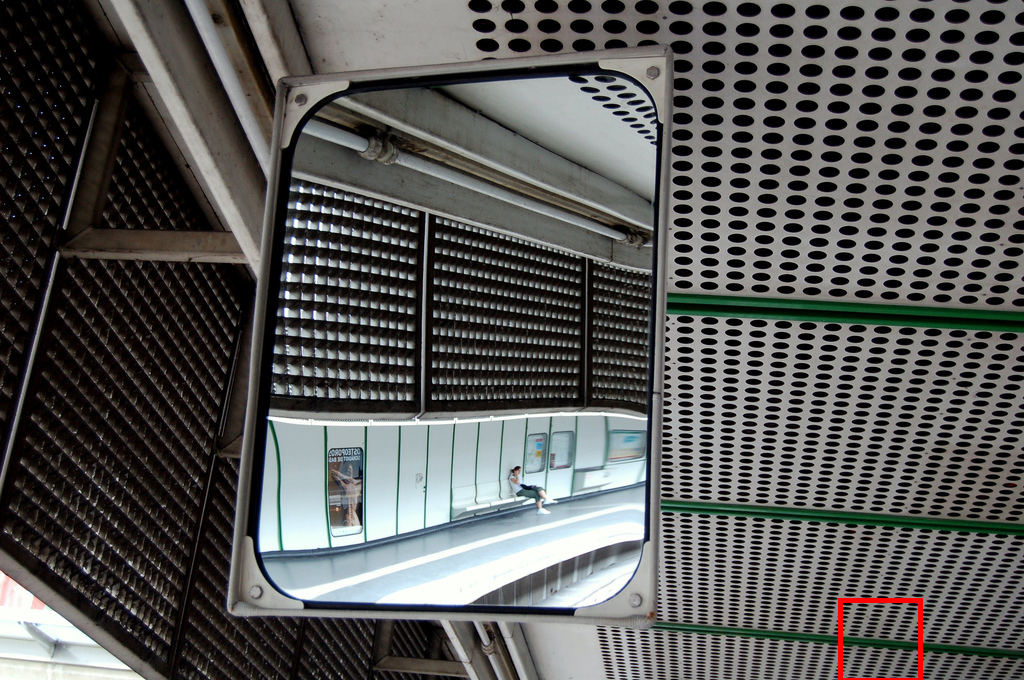}
    \caption{HR\wt{$^\dagger$}}
\end{subfigure} \hspace*{-0.4em}
\begin{subfigure}[b]{0.12\textwidth}
    \centering
    \includegraphics[width=\textwidth, interpolate=false]{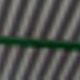}
    \caption{Bicubic\wt{$^\dagger$}}
\end{subfigure} \hspace*{-0.4em}
\begin{subfigure}[b]{0.12\textwidth}
    \centering
    \includegraphics[width=\textwidth, interpolate=false]{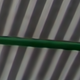}
    \caption{RCAN\wt{$^\dagger$}}
\end{subfigure} \hspace*{-0.4em}
\begin{subfigure}[b]{0.12\textwidth}
    \centering
\includegraphics[width=\textwidth, interpolate=false]{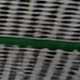}
    \caption{CAR\wt{$^\dagger$}}
\end{subfigure} \hspace*{-0.4em}
\begin{subfigure}[b]{0.12\textwidth}
    \centering
\includegraphics[height=\textwidth, interpolate=false]{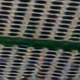}
    \caption{\textbf{IRN$^\dagger$}}
    \end{subfigure} \hspace*{-0.4em}
\begin{subfigure}[b]{0.12\textwidth}
    \centering
    \includegraphics[width=\textwidth, interpolate=false]{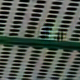}
    \caption{HCFlow\wt{$^\dagger$}}
\end{subfigure} \hspace*{-0.4em}
\begin{subfigure}[b]{0.12\textwidth}
    \centering
    \includegraphics[width=\textwidth, interpolate=false]{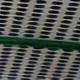}
    \caption{DLV-IRN\wt{$^\dagger$}}
\end{subfigure} \hspace*{-0.4em}
\begin{subfigure}[b]{0.12\textwidth}
    \centering
    \includegraphics[width=\textwidth, interpolate=false]{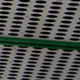}
    \caption{GT\wt{$^\dagger$}}
\end{subfigure} \hspace*{-0.4em}

\end{center}

\caption{Visual comparisons of upscaling the $\times 4$ downscaled images. }

\label{fig:hr}
\end{figure*}

\noindent\textbf{Dual latent variable enhancement of IRN.}
For quantitative comparison in restored HR image quality in Table \ref{tab:hr}, IRN is used as the primary baseline model for dual latent variable enhancement.
The Type I category includes image SR models optimized for upscaling only.  They are separately listed from bidirectional models in Type II category for fair comparison,
as the latter ones have the advantage of jointly optimizing downscaling and upscaling which is demonstrated in the big difference in PSNR and SSIM between two categories.
For Type II models, our enhanced DLV-IRN is consistently better than other INN based models which has only the default upscaling latent variable $\mathbf{z}$, including the retrained
IRN$^\dagger$ which is equivalent with DLV-IRN in model size.  Comparing to the latest FGRN which does not have latent variables so that our enhancement is not applicable,
our DLV-IRN is still better in large test sets like BSD100, Urban100 and DIV2K overall, while trailing behind in smaller test sets.
From visual examples shown in Fig. \ref{fig:hr}, it is clear that our DLV-IRN is capable of restoring high frequency details more precisely.

\begin{figure*}[t!]
\captionsetup[subfigure]{font=footnotesize, labelformat=empty}
\begin{center} \hspace*{1.5em}
\begin{subfigure}[b]{0.0655\textwidth}
    \centering
    \includegraphics[width=\textwidth, height=2\textwidth, interpolate=false]{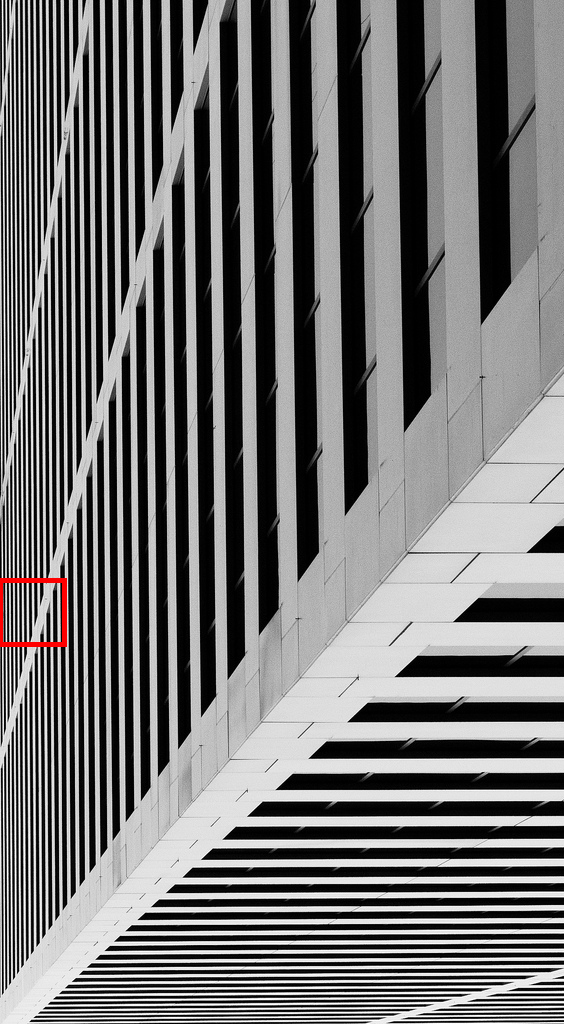}
    \end{subfigure} \hspace*{1.5em}
\begin{subfigure}[b]{0.13\textwidth}
    \centering
    \includegraphics[width=\textwidth, interpolate=false]{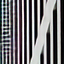}
\end{subfigure}  
\begin{subfigure}[b]{0.13\textwidth}
    \centering
    \includegraphics[width=\textwidth, interpolate=false]{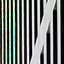}
\end{subfigure}  
\begin{subfigure}[b]{0.13\textwidth}
    \centering
    \includegraphics[width=\textwidth, interpolate=false]{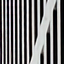}
\end{subfigure} 
\begin{subfigure}[b]{0.13\textwidth}
    \centering
    \includegraphics[width=\textwidth, interpolate=false]{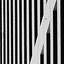}
\end{subfigure} 

\begin{subfigure}[b]{0.13\textwidth}
    \centering
    \includegraphics[width=\textwidth, height=\textwidth, interpolate=false]{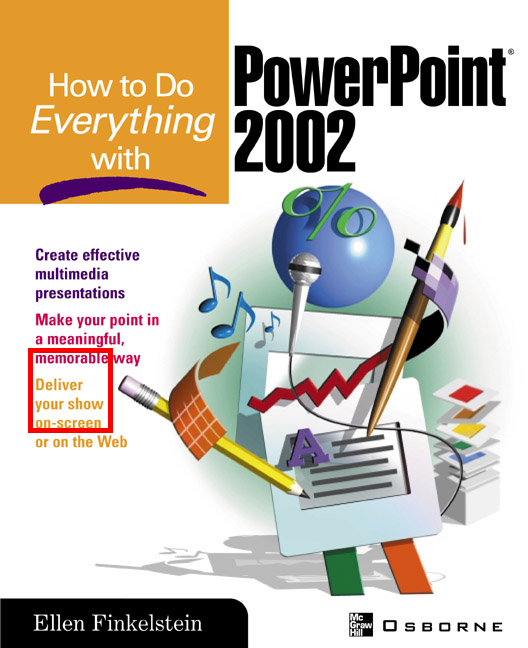}
    \caption{HR}
    \end{subfigure} 
\begin{subfigure}[b]{0.13\textwidth}
    \centering
    \includegraphics[width=\textwidth, interpolate=false]{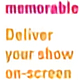}
    \caption{HCFlow}
\end{subfigure} 
\begin{subfigure}[b]{0.13\textwidth}
    \centering
    \includegraphics[width=\textwidth, interpolate=false]{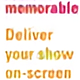}
    \caption{HCFlow+}
\end{subfigure} 
\begin{subfigure}[b]{0.13\textwidth}
    \centering
    \includegraphics[width=\textwidth, interpolate=false]{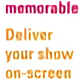}
    \caption{DLV-HCFlow}
\end{subfigure} 
\begin{subfigure}[b]{0.13\textwidth}
    \centering
    \includegraphics[width=\textwidth, interpolate=false]{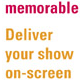}
    \caption{GT}
\end{subfigure} 
\end{center}

\caption{Visual comparisons of image rescaling ($\times 4$) by the family of HCFlow which are trained on the DIV2K set.}

\label{fig:hcflow2}
\end{figure*}

\begin{figure*}[t!]
\begin{center}

  \begin{subfigure}[b]{0.075\textwidth}
    \centering
      \includegraphics[width=\textwidth, interpolate=false]{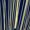}
  \end{subfigure} \hspace*{-0.4em}
  \begin{subfigure}[b]{0.075\textwidth}
    \centering
      \includegraphics[width=\textwidth, interpolate=false]{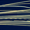}
  \end{subfigure} \hspace*{-0.4em}
  \begin{subfigure}[b]{0.075\textwidth}
    \centering
      \includegraphics[width=\textwidth, interpolate=false]{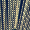}
  \end{subfigure} \hspace*{-0.4em}
  \begin{subfigure}[b]{0.075\textwidth}
    \centering
      \includegraphics[width=\textwidth, interpolate=false]{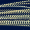}
  \end{subfigure} \hspace*{-0.4em}
  \begin{subfigure}[b]{0.075\textwidth}
    \centering
      \includegraphics[width=\textwidth, interpolate=false]{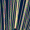}
  \end{subfigure} \hspace*{-0.4em}
  \begin{subfigure}[b]{0.075\textwidth}
    \centering
      \includegraphics[width=\textwidth, interpolate=false]{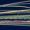}
  \end{subfigure} \hspace*{-0.4em}
  \begin{subfigure}[b]{0.075\textwidth}
    \centering
      \includegraphics[width=\textwidth, interpolate=false]{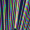}
  \end{subfigure} \hspace*{-0.4em}
  \begin{subfigure}[b]{0.075\textwidth}
    \centering
      \includegraphics[width=\textwidth, interpolate=false]{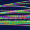}
  \end{subfigure} \hspace*{-0.4em}
  \begin{subfigure}[b]{0.075\textwidth}
    \centering
      \includegraphics[width=\textwidth, interpolate=false]{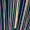}
  \end{subfigure} \hspace*{-0.4em}
  \begin{subfigure}[b]{0.075\textwidth}
    \centering
      \includegraphics[width=\textwidth, interpolate=false]{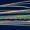}
  \end{subfigure} \hspace*{-0.4em}
  \begin{subfigure}[b]{0.075\textwidth}
    \centering
      \includegraphics[width=\textwidth, interpolate=false]{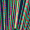}
  \end{subfigure} \hspace*{-0.4em}
  \begin{subfigure}[b]{0.075\textwidth}
    \centering
      \includegraphics[width=\textwidth, interpolate=false]{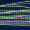}
  \end{subfigure}
  
  \begin{subfigure}[b]{0.152\textwidth}
    \centering
      \includegraphics[width=\textwidth, interpolate=false]{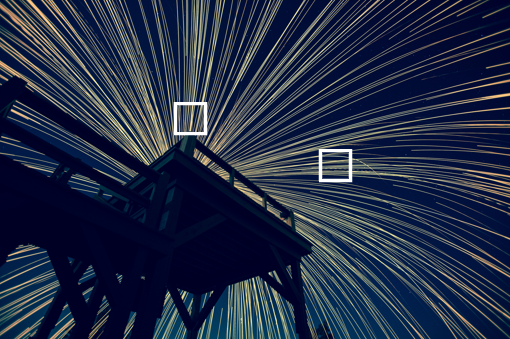}
      \caption{Bicubic\wt{$^\dagger$}}
  \end{subfigure} \hspace*{-0.41em}
  \begin{subfigure}[b]{0.152\textwidth}
    \centering
      \includegraphics[width=\textwidth, interpolate=false]{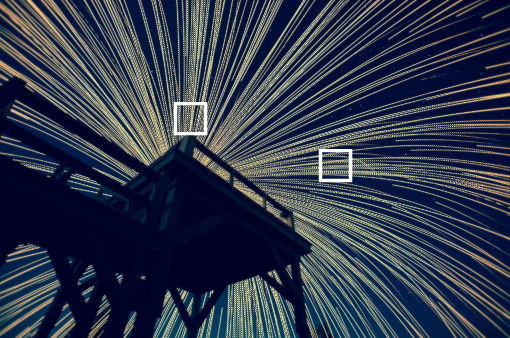}
      \caption{\textbf{CAR}\wt{$^\dagger$}}
  \end{subfigure} \hspace*{-0.41em}
  \begin{subfigure}[b]{0.152\textwidth}
    \centering
      \includegraphics[width=\textwidth, interpolate=false]{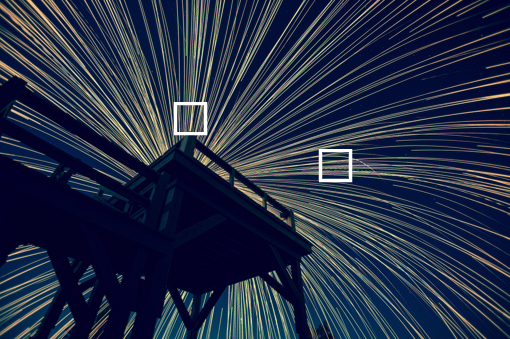}
      \caption{\textbf{IRN$^\dagger$}}
  \end{subfigure} \hspace*{-0.41em}
  \begin{subfigure}[b]{0.152\textwidth}
    \centering
      \includegraphics[width=\textwidth, interpolate=false]{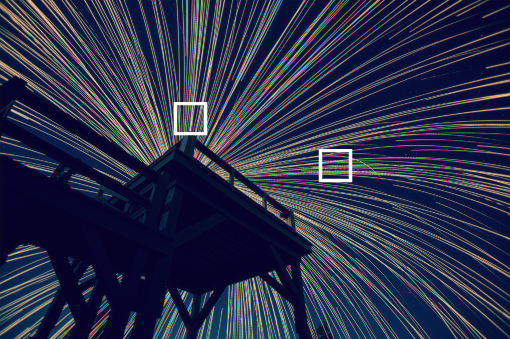}
      \caption{\textbf{HCFlow$^\dagger$}}
  \end{subfigure} \hspace*{-0.41em}
  \begin{subfigure}[b]{0.152\textwidth}
    \centering
      \includegraphics[width=\textwidth, interpolate=false]{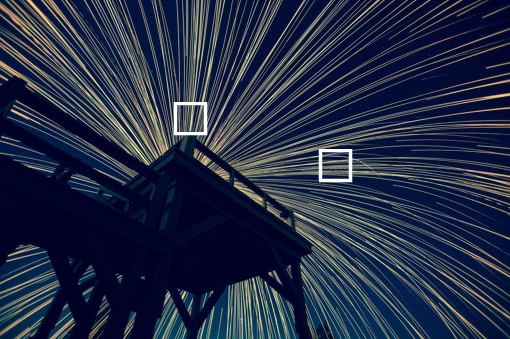}
      \caption{DLV-IRN\wt{$^\dagger$}}
  \end{subfigure} \hspace*{-0.41em}
  \begin{subfigure}[b]{0.152\textwidth}
    \centering
      \includegraphics[width=\textwidth, interpolate=false]{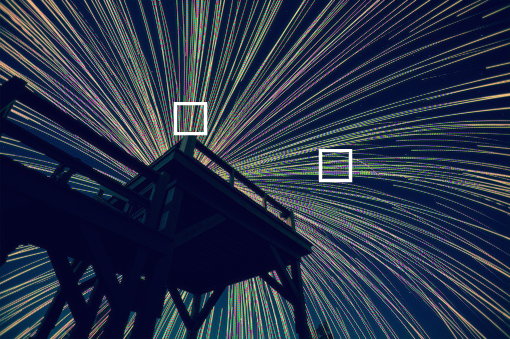}
      \caption{DLV-HCFlow\wt{$^\dagger$}}
  \end{subfigure}
 \end{center}
\caption{Visual examples of downscaled ($\times4$) LR images, selected to demonstrate visual artifacts in worst cases.}
\label{fig:lr}
\end{figure*}

\noindent\textbf{Dual latent variable enhancement of HCFlow.}
The dual latent variable enhancement is also applied to HCFlow to train a new model DLV-HCFlow.  In addition to validating the general effectiveness of DLV-enhancement on different
models, this experiment is summarized separately from the main one for a couple of reasons.
First, the baseline HCFlow is trained from DF2K, a much larger dataset than the DIV2K as used in IRN.  To study the effects of different training set sizes, all three models, including
HCFlow, its variant HCFlow$^\dagger$ with increased model depth and the enhanced DLV-HCFlow, are all trained with DF2K and DIV2K respectively.  As shown in Table \ref{tab:hcflow},
for certain training set, there is marginal improvement from HCFlow to HCFlow$^\dagger$ due to increased model depth.
For comparison between the two training sets, it shows that results from DIV2K are at least as good as those from DF2K.  This is in contrast to what is commonly observed from other image restoration models
like ESRGAN \cite{wang_eccv_2018} where increased training set size lead to performance improvement in general.
Secondly, it is found out that HCFlow, and accordingly HCFlow$^\dagger$ and DLV-HCFlow, uses a smaller weight for guidance loss $L_g$, which leads to poor image quality in
downscaled LR images, as demonstrated later in Fig. \ref{fig:lr}.
In other words, the performance improvement in HCFlow over IRN for upscaled HR images is accompanied with sacrifices in LR image quality.
Nevertheless, the DLV enhancement is shown to be effective for HCFlow too.  For DLV-HCFlow, there are additional improvements across different test sets
comparing to HCFlow$^\dagger$, although both are similar in model size.  For the challenging case of Urban100, the additional increase in PSNR is as large as 0.31.
From visual examples in Fig. \ref{fig:hcflow2}, it is also clear that DLV-HCFlow restores fine details better and has less visible artifacts.

\begin{table}[t!]
    \centering
    \caption{Upscaled ($\times 4$) HR results with the best ones in bold.}
    \vspace{8pt}
    \label{tab:hcflow}
    \begin{NiceTabular}{c|c|c|c|c|c|c} \hline
        \multirow{2}*{Methods} & \multirow{2}*{Param} & \multicolumn{5}{c}{\tabularnote{PSNR$^{1,2}$ are from models trained on DF2K and DIV2K dataset respectively}PSNR$^1$/PSNR$^2$} \\ \cline{3-7}
        & & Set5 & Set14 & BSD100 & Urban100 & DIV2K \\ \hline\hline
        HCFlow \cite{liang2021hierarchical} & 4.4M & 36.29/36.24 & 33.02/33.10 & 31.74/31.71 & 31.62/31.96 & 35.23/35.27 \\
        HCFlow$^\dagger$ & 4.93M & 36.25/36.29 & 32.96/33.14 & 31.74/31.78 & 31.68/32.13 & 35.25/35.36 \\
        DLV-HCFlow & 4.87M & \textbf{36.40}/36.38 & 33.30/\textbf{33.33} & 31.82/\textbf{31.83} & 31.99/\textbf{32.41} & 35.39/\textbf{35.43} \\ \hline
    \end{NiceTabular}
\end{table}

\begin{table*}[t!]
    \centering
    \footnotesize
    \caption{Quantitative results of LR ($\times 4$) image quality by different downscaling methods.}
    \label{tab:lr}
    \setlength{\tabcolsep}{2.5pt}
    \resizebox{\textwidth}{!}{%
    \begin{NiceTabular}{c|c|ccc|c|ccc|c|ccc|c|ccc|c|ccc} \hline
        \multirow{3}*{Methods} & \multicolumn{4}{c|}{Set5} & \multicolumn{4}{c|}{Set14} & \multicolumn{4}{c|}{BSD100} & \multicolumn{4}{c|}{Urban100} & \multicolumn{4}{c}{DIV2K} \\ \cline{2-21}
        & Y & \multicolumn{3}{c|}{RGB} & Y & \multicolumn{3}{c|}{RGB} & Y & \multicolumn{3}{c|}{RGB} & Y & \multicolumn{3}{c|}{RGB} & Y & \multicolumn{3}{c}{RGB} \\ \cline{2-21}
        & SSIM$\uparrow$ & SSIM$\uparrow$ & NIQE$\downarrow$ & PIQE$\downarrow$ & SSIM$\uparrow$ & SSIM$\uparrow$ & NIQE$\downarrow$ & PIQE$\downarrow$ & SSIM$\uparrow$ & SSIM$\uparrow$ & NIQE$\downarrow$ & PIQE$\downarrow$ & SSIM$\uparrow$ & SSIM$\uparrow$ & NIQE$\downarrow$ & PIQE$\downarrow$ & SSIM$\uparrow$ & SSIM$\uparrow$ & NIQE$\downarrow$ & PIQE$\downarrow$ \\ \hline\hline
        Bicubic & - & - & 18.875 & 52.043 & - & - & 17.917 & 44.927 & - & - & 18.878 & 42.072 & - & - & \blue{17.214} & 46.667 & - & - & \red{3.979} & 38.322 \\
        CAR & 0.9628 & 0.9532 & \red{18.873} & 59.782 & 0.9358 & 0.9303 & 17.936 & 46.005 & 0.9226 & 0.9160 & \red{18.878} & 42.263 & 0.9196 & 0.9146 & 22.731 & 48.959 & 0.9460 & 0.9404 & 5.549 & 38.127 \\
        HCFlow$^\dagger$ & 0.9846 & 0.9200 & 18.875 & 49.174 & 0.9766 & 0.8789 & 17.943 & 41.457 & 0.9751 & 0.8605 & 18.879 & 40.887 & 0.9704 & 0.8546 & 18.895 & 42.729 & 0.9792 & 0.8786 & 5.174 & 33.319 \\
        DLV-HCFlow & 0.9797 & 0.9293 & 18.875 & \blue{43.697} & 0.9609 & 0.8877 & 17.941 & 42.514 & 0.9562 & 0.8650 & 18.879 & 40.791 & 0.9549 & 0.8579 & 18.010 & 42.058 & 0.9670 & 0.8857 & 5.194 & 33.441 \\
        IRN$^\dagger$ & \blue{0.9960} & \blue{0.9823} & 18.875 & 44.920  & \blue{0.9926} & \blue{0.9705} & \blue{17.896} & \blue{39.512} & \blue{0.9922} & \blue{0.9666} & 18.879 & \blue{36.123} & \blue{0.9917} & \blue{0.9684} & 17.527 & \blue{39.838} & \blue{0.9930} & \blue{0.9696} & 4.235 & \blue{30.644} \\
        DLV-IRN & \red{0.9961} & \red{0.9833} & \blue{18.875} & 48.822  & \red{0.9937} & \red{0.9730} & \red{17.890} & 40.027 & \red{0.9924} & \red{0.9670} & 18.879 & 36.855 & \red{0.9930} & \red{0.9701} & 17.613 & 40.486 & \red{0.9936} & \red{0.9712} & \blue{4.211} & 30.864 \\ 
        FGRN & - & - & 18.876 & \red{40.627} & - & - & 17.952 & \red{35.464} & - & - & \blue{18.878} & \red{34.282} & - & - & \red{17.132} & \red{36.991} & - & - & 4.358 & \red{27.314} \\ \hline
    \end{NiceTabular}}
\end{table*}


\noindent\textbf{Quality assessment of downscaled LR images.}
Although there is no GT reference for downscaled LR in this study, it is important to generate LR images which are visually consistent with bicubically downsampled LR and free of obvious visual
artifacts.  Considering this, LR outputs from different methods are compared in both full reference image quality assessment like SSIM and no reference ones like NIQE and PIQE.
While the majority of generated LR images are of high qualities, a worst-case example from DIV2K validation set is selected here to demonstrate potential severity of visual artifacts.
As shown in Fig. \ref{fig:lr}, when viewed in its native resolution, generated LR from IRN$^\dagger$ and DLV-IRN look very similar to bicubic reference without obvious artifacts.  For CAR, it is visibly brighter than normal overall.
For the HCFlow family, false color and Moiré-like artifacts are noticeable even without zoomed-in.
In magnified views, both false color and Moiré-like artifacts are very obvious, and the false color ones are more severe in general.
Considering this, the SSIM metric is calculated twice for Y channel only and RGB channels respectively.  As shown in Table \ref{tab:lr}, the difference between two SSIM values
is minimum for CAR, which is consistent with visual examples in Fig. \ref{fig:lr} where no false color artifacts noticed for CAR.  In contrast, this difference is very large for the HCFlow family,
as their false color effects are the worst as demonstrated in Fig. \ref{fig:lr}.  Results of the HCFlow family are also worse than the IRN ones in both NIQE and PIQE.
For DLV-IRN and DLV-HCFlow, the difference between them and the corresponding IRN$^\dagger$ and HCFlow$^\dagger$ are minor, as demonstrated in both qualitative results in Fig. \ref{fig:lr}
and quantitative values in Table \ref{tab:lr}.  In summary, the performance enhancement in HR restoration brought by dual latent variables is achieved without sacrificing image quality
in downscaled LR.  In contrast, the improvement of HCFlow over IRN is accompanied with significant degradation in LR.  While FGRN\footnote{NIQE and PIQE are quoted from  \cite{li2021approaching} while others are computed in MATLAB following \url{https://www.mathworks.com/help/images/image-quality.html}} is the best in PIQE, there are no SSIM values and downscale images
available for more comprehensive comparison.

\begin{table*}[t!]
    \centering
	\caption{Image hiding results of different methods on benchmark datasets.}
	\label{tab:hinet}
	\resizebox{\textwidth}{!}{%
    \begin{tabular}{c|c|cccc|cccc|cccc} \hline\hline
        \multirow{3}*{Methods} & \multirow{3}*{Param} & \multicolumn{12}{c}{Cover/Stego image pair} \\ \cline{3-14}
         & & \multicolumn{4}{c|}{DIV2K} & \multicolumn{4}{c|}{COCO} & \multicolumn{4}{c}{ImageNet} \\ \cline{3-14}
         & & PSNR(dB)$\uparrow$ & SSIM$\uparrow$ & MAE$\downarrow$ & RMSE$\downarrow$ & PSNR(dB)$\uparrow$ & SSIM$\uparrow$ & MAE$\downarrow$ & RMSE$\downarrow$ & PSNR(dB)$\uparrow$ & SSIM$\uparrow$ & MAE$\downarrow$ & RMSE$\downarrow$ \\ \hline
         4bit-LSB & - & 33.19 & 0.9453 & 6.90 & 7.95 & 33.79 & 0.9479 & 7.31 & 9.12 & 33.68 & 0.9401 & 6.46 & 8.48 \\ 
         HiDDeN \cite{zhu2018hidden} & - & 35.21 & 0.9691 & 6.98 & 6.82 & 36.71 & 0.9876 & 6.58 & 8.73 & 34.79 & 0.9380 & 6.12 & 7.33 \\
         Baluja \cite{baluja2017hiding} & 42.6M & 41.95 & 0.9838 & 2.48 & 3.44 & 39.15 & 0.9770 & 3.43 & 4.83 & 39.19 & 0.9769 & 3.49 & 4.85 \\
         HiNet \cite{jing2021hinet} & 4.1M & 44.94 & 0.9864 & 2.07 & 2.84 & 41.73 & 0.9776 & 3.03 & 4.22 & 41.54 & 0.9759 & 3.15 & 4.33 \\
         HiNet$^\dagger$ & 4.5M & 45.19 & 0.9868 & 1.97 & 2.73 & 42.00 & 0.9786 & 2.92 & 4.07 & 41.85 & 0.9771 & 3.03 & 4.17 \\ 
         DLV-HiNet & 4.5M & \textbf{46.65} & \textbf{0.9902} & \textbf{1.80} & \textbf{2.50} & \textbf{42.93} & \textbf{0.9816} & \textbf{2.73} & \textbf{3.81} & \textbf{42.74} & \textbf{0.9800} & \textbf{2.85} & \textbf{3.91} \\ \hline\hline
         \multirow{3}*{Methods} & \multirow{3}*{Param} & \multicolumn{12}{c}{Secret/Recovery image pair} \\ \cline{3-14}
         & & \multicolumn{4}{c|}{DIV2K} & \multicolumn{4}{c|}{COCO} & \multicolumn{4}{c}{ImageNet} \\ \cline{3-14}
         & & PSNR(dB)$\uparrow$ & SSIM$\uparrow$ & MAE$\downarrow$ & RMSE$\downarrow$ & PSNR(dB)$\uparrow$ & SSIM$\uparrow$ & MAE$\downarrow$ & RMSE$\downarrow$ & PSNR(dB)$\uparrow$ & SSIM$\uparrow$ & MAE$\downarrow$ & RMSE$\downarrow$ \\ \hline
         4bit-LSB & - & 30.81 & 0.9020 & 8.96 & 8.01 & 32.04 & 0.9127 & 7.61 & 9.59 & 31.26 & 0.9033 & 7.71 & 9.76 \\ 
         HiDDeN \cite{zhu2018hidden} & - & 36.43 & 0.9696 & 6.02 & 5.50 & 37.68 & 0.9845 & 4.72 & 6.33 & 35.70 & 0.9601 & 4.57 & 6.92 \\
         Baluja \cite{baluja2017hiding} & 42.6M & 40.32 & 0.9776 & 2.62 & 3.53 & 38.00 & 0.9705 & 3.62 & 5.02 & 37.91 & 0.9697 & 3.72 & 5.12 \\
         HiNet \cite{jing2021hinet} & 4.1M & 49.32 & 0.9977 & 0.92 & 1.31 & 46.65 & 0.9964 & 1.45 & 2.12 & 46.49 & 0.9960 & 1.54 & 2.21 \\
         HiNet$^\dagger$ & 4.5M &49.67 & 0.9977 & 0.88 & 1.27 & 46.95 & 0.9965 & 1.41 & 2.07 & 46.78 & 0.9961 & 1.51 & 2.17 \\
         DLV-HiNet & 4.5M & \textbf{50.12} & \textbf{0.9979} & \textbf{0.83} & \textbf{1.21} & \textbf{47.51} & \textbf{0.9968} & \textbf{1.32} & \textbf{1.93} & \textbf{47.36} & \textbf{0.9965} & \textbf{1.40} & \textbf{2.01} \\ \hline
    \end{tabular}}
\end{table*}

\begin{figure*}[t!]
\begin{center}
\hspace*{0.009em}
\rotatebox[origin=l]{90}{\makebox[0.07\textwidth][r]{Cover}} 
\begin{subfigure}[b]{0.12\textwidth}
    \centering
    \includegraphics[width=\textwidth, interpolate=false]{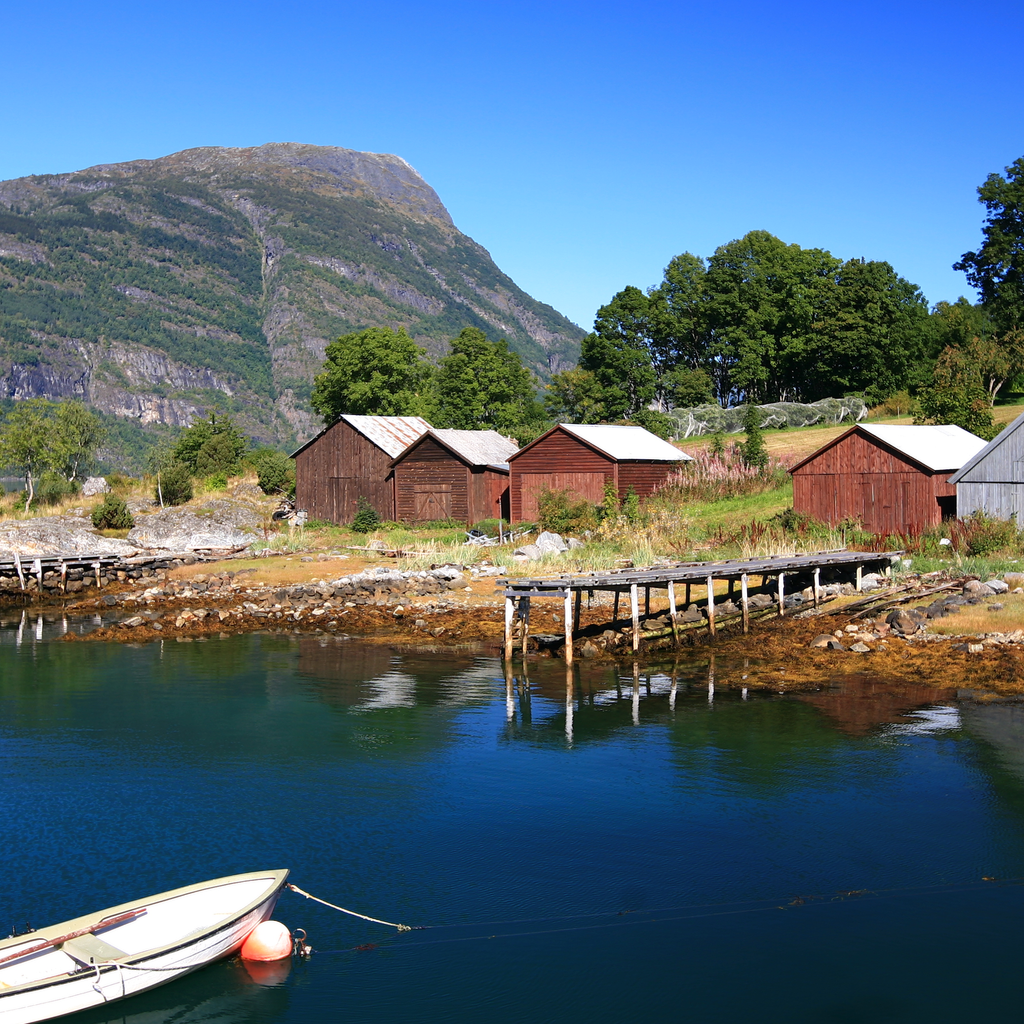}
    \end{subfigure} 
\begin{subfigure}[b]{0.12\textwidth}
    \centering
    \includegraphics[width=\textwidth, interpolate=false]{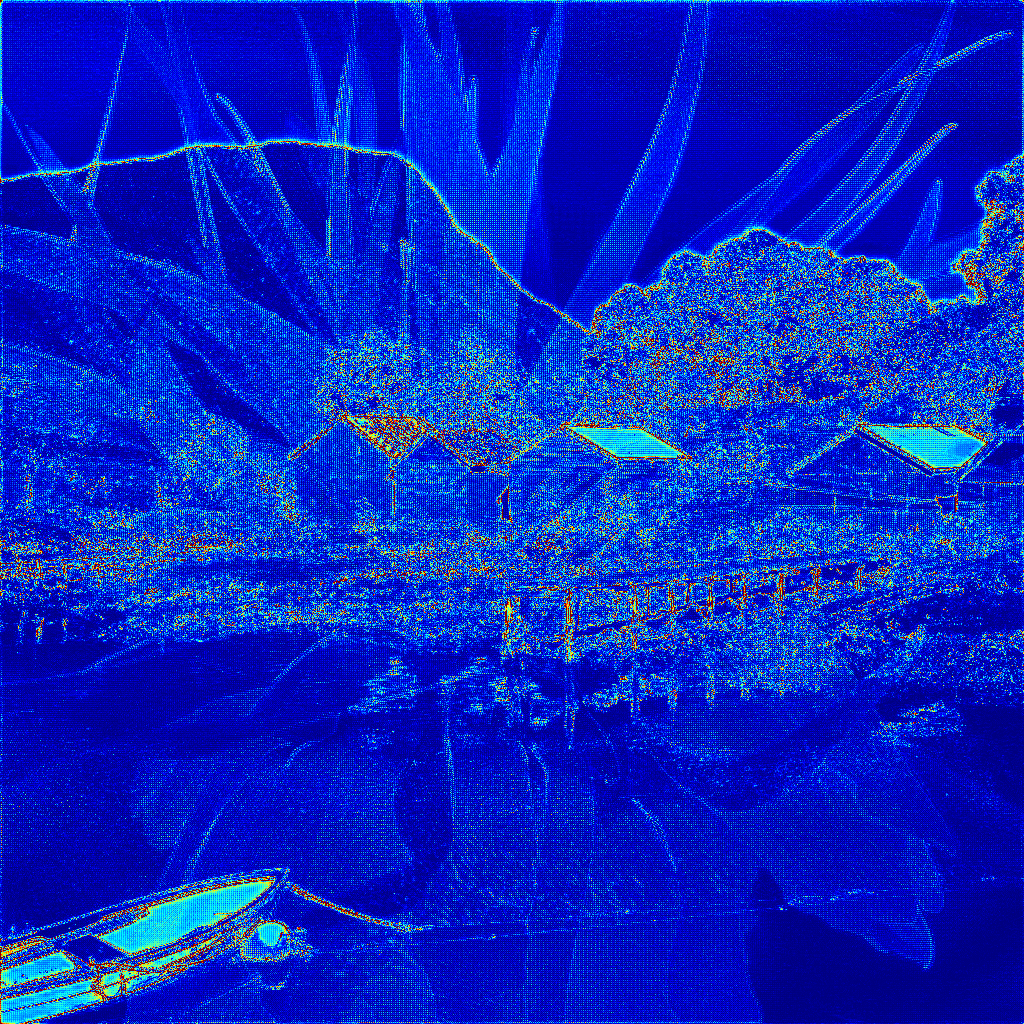}
\end{subfigure} 
\begin{subfigure}[b]{0.12\textwidth}
    \centering
    \includegraphics[width=\textwidth, interpolate=false]{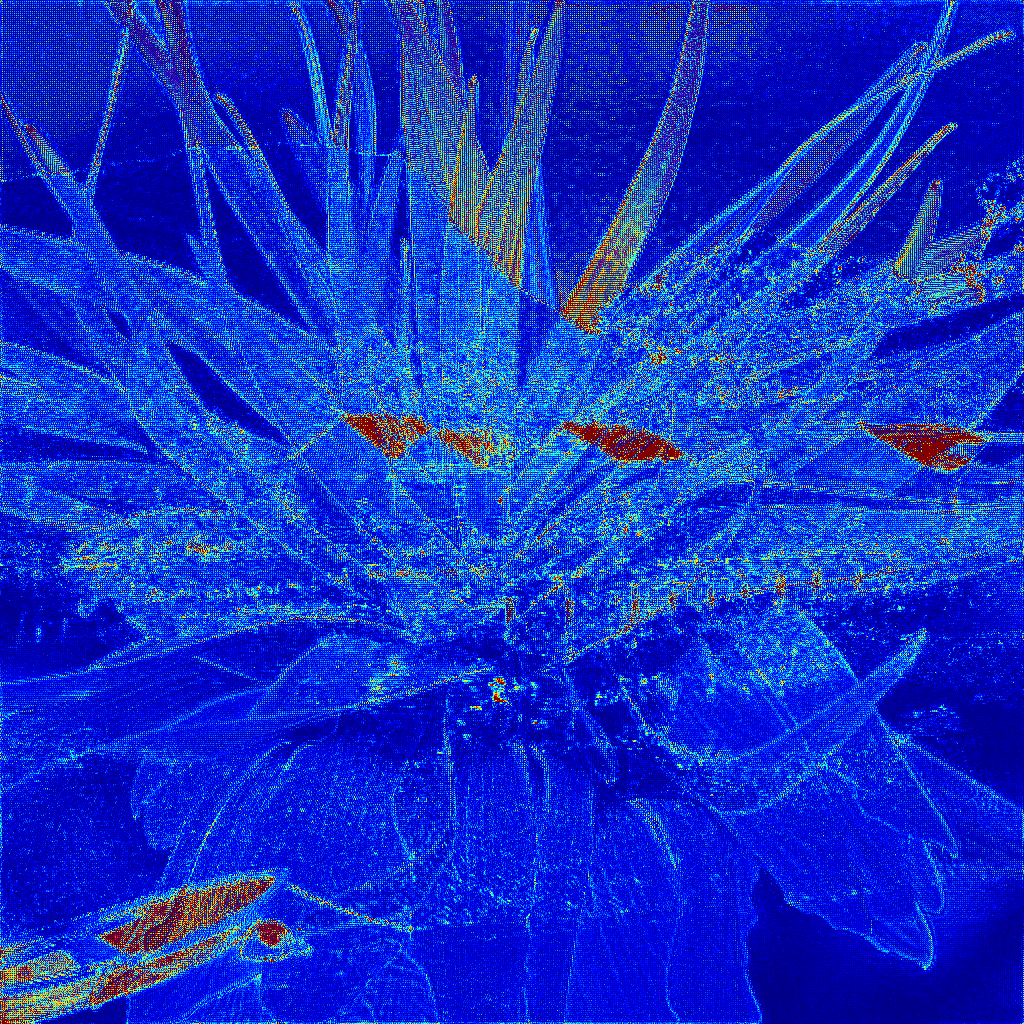}
\end{subfigure} 
\begin{subfigure}[b]{0.12\textwidth}
    \centering
    \includegraphics[width=\textwidth, interpolate=false]{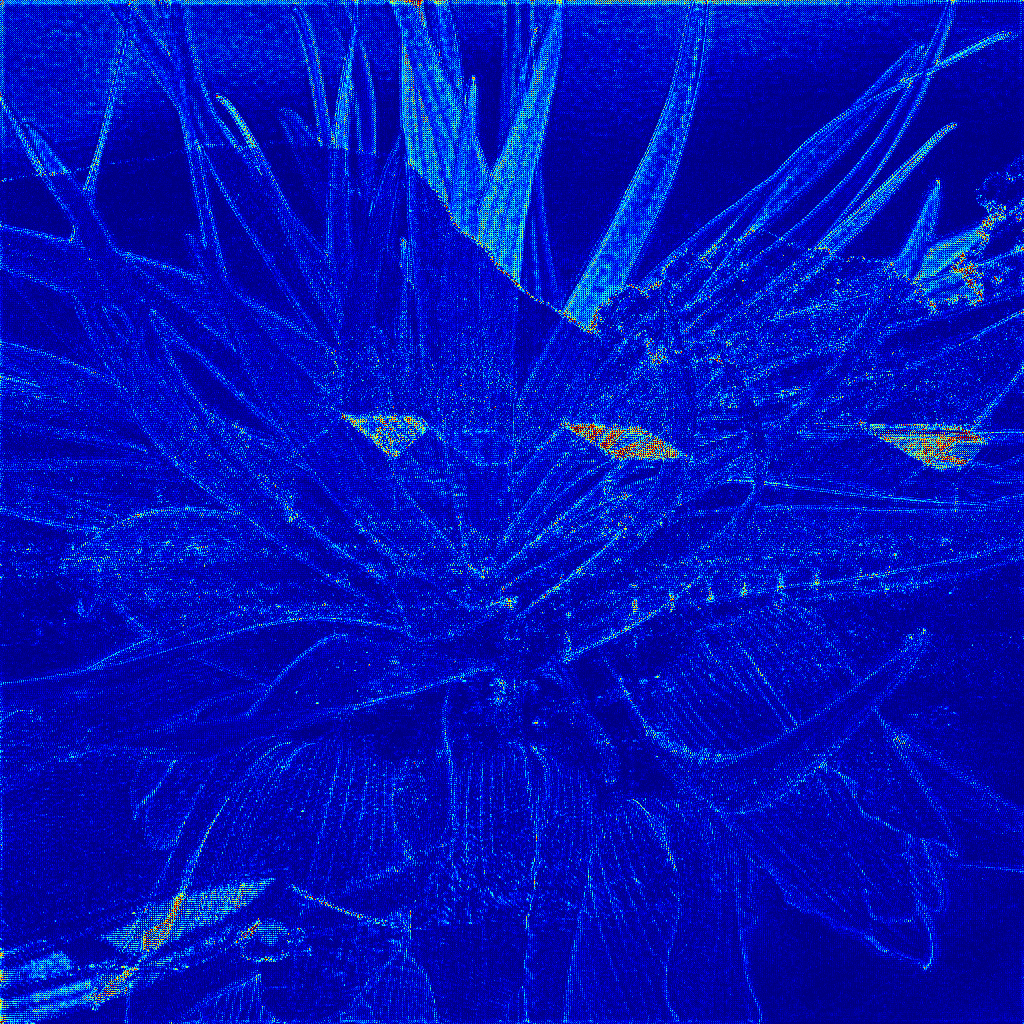}
\end{subfigure} 
\begin{subfigure}[b]{0.12\textwidth}
    \centering
    \includegraphics[width=\textwidth, interpolate=false]{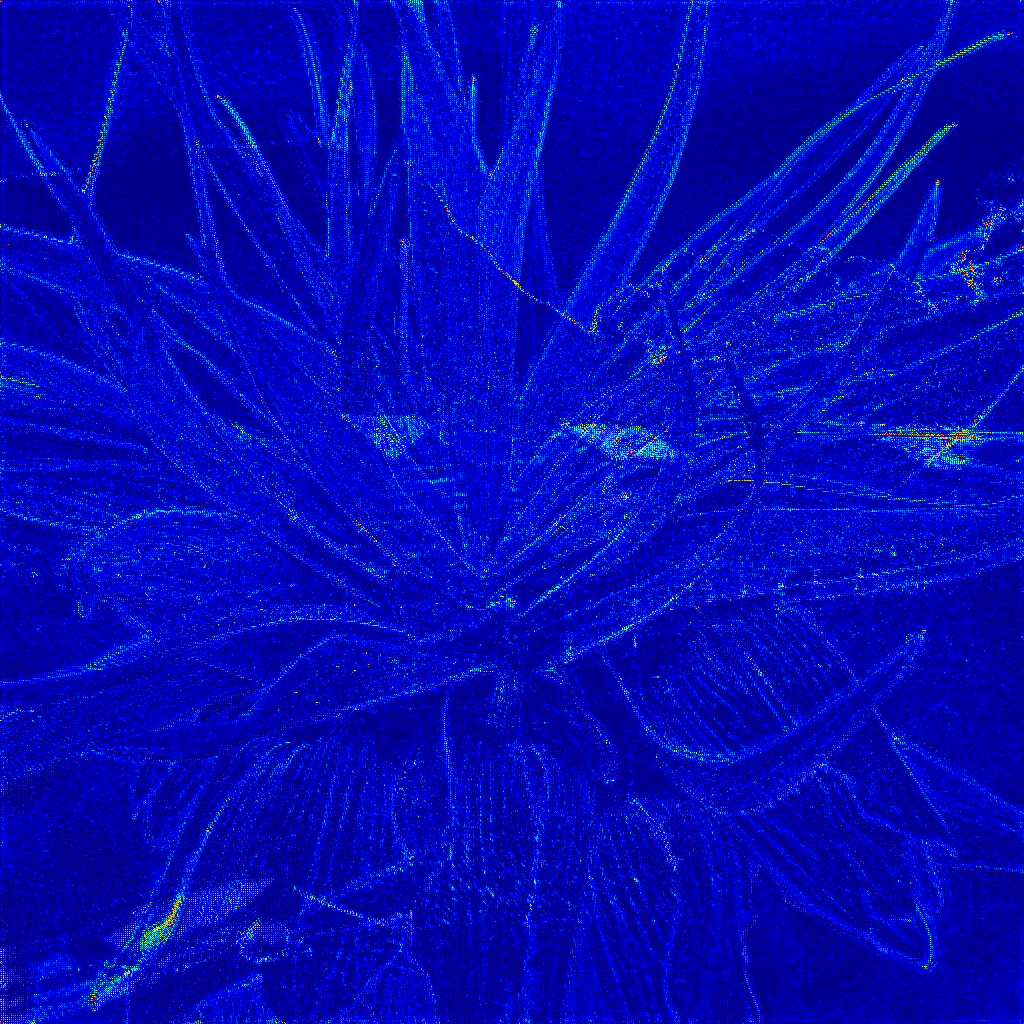}
\end{subfigure} 
\begin{subfigure}[b]{0.12\textwidth}
    \centering
    \includegraphics[width=\textwidth, interpolate=false]{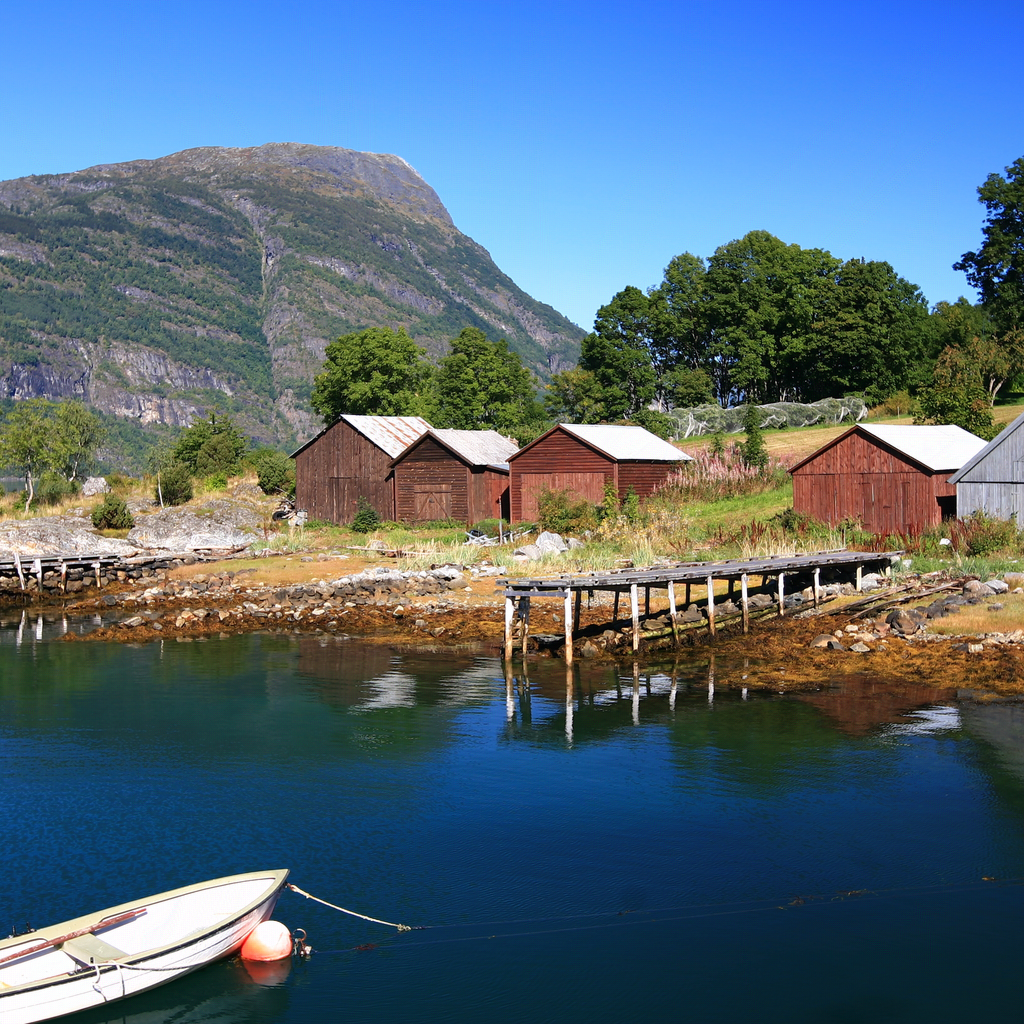}
\end{subfigure} 
\rotatebox[origin=l]{90}{\makebox[0.07\textwidth][r]{Stego}}

\rotatebox[origin=l]{90}{\makebox[0.11\textwidth][r]{Secret}}
\begin{subfigure}[b]{0.12\textwidth}
    \centering
    \includegraphics[width=\textwidth, interpolate=false]{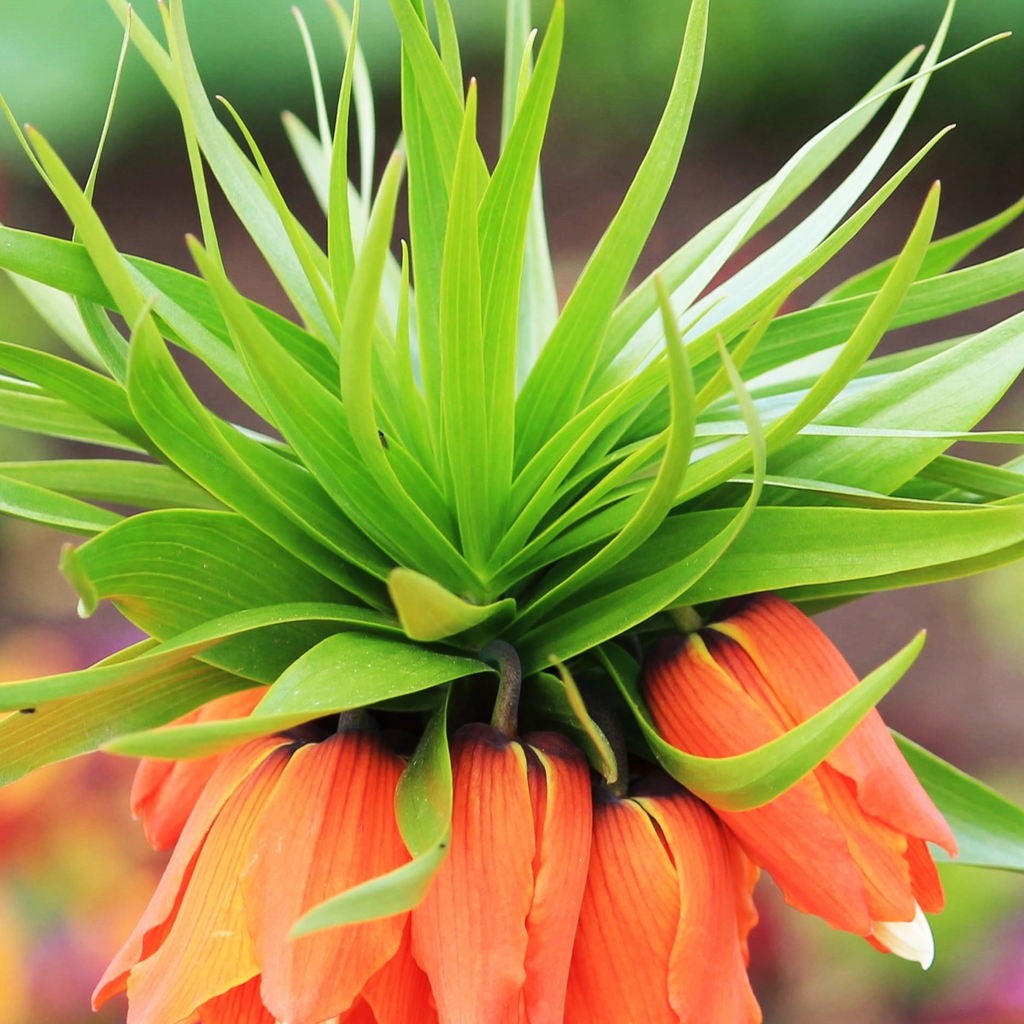}
    \caption{GT}
\end{subfigure} 
\begin{subfigure}[b]{0.12\textwidth}
    \centering
    \includegraphics[width=\textwidth, interpolate=false]{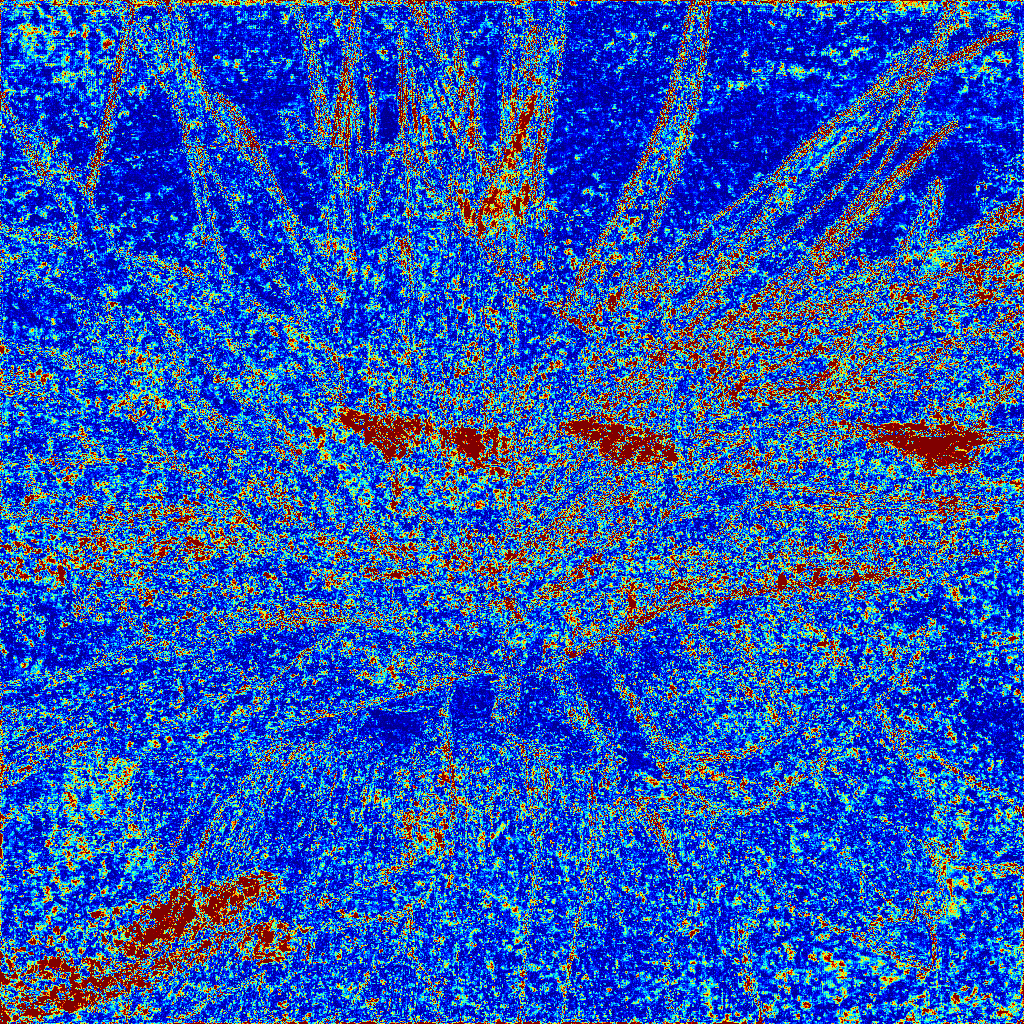}
    \caption{Baluja}
\end{subfigure} 
\begin{subfigure}[b]{0.12\textwidth}
    \centering
    \includegraphics[width=\textwidth, interpolate=false]{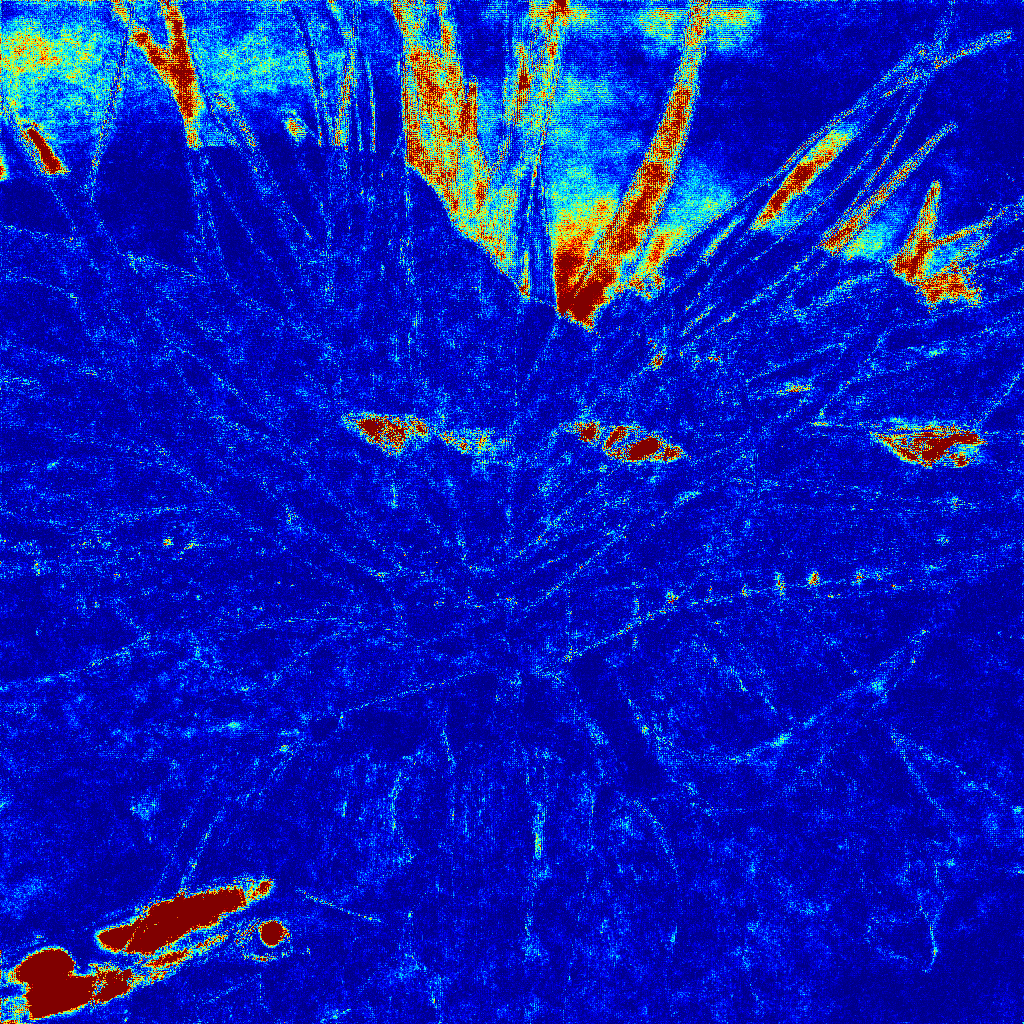}
    \caption{HiNet}
\end{subfigure} 
\begin{subfigure}[b]{0.12\textwidth}
    \centering
    \includegraphics[width=\textwidth, interpolate=false]{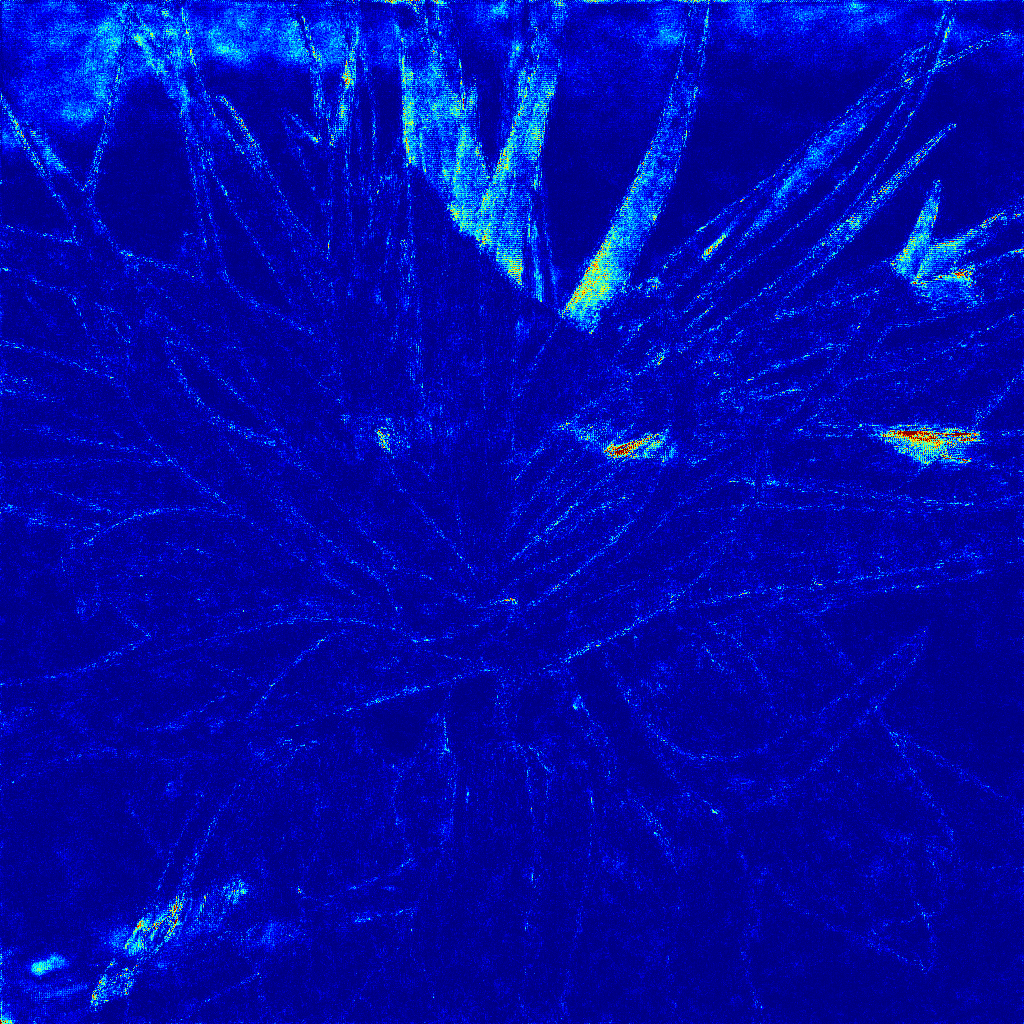}
    \caption{HiNet+}
\end{subfigure} 
\begin{subfigure}[b]{0.12\textwidth}
    \centering
    \includegraphics[width=\textwidth, interpolate=false]{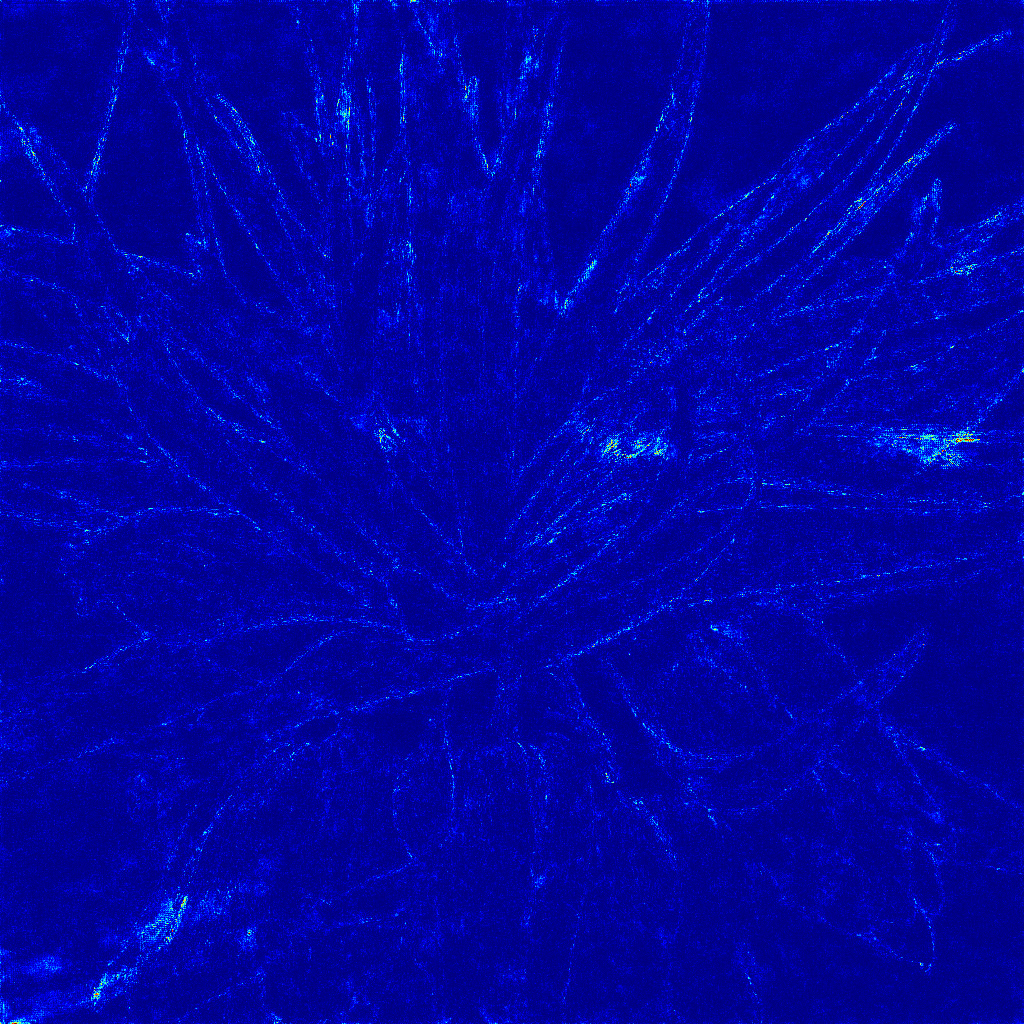}
    \caption{DLV-HiNet}
\end{subfigure} 
\begin{subfigure}[b]{0.12\textwidth}
    \centering
    \includegraphics[width=\textwidth, interpolate=false]{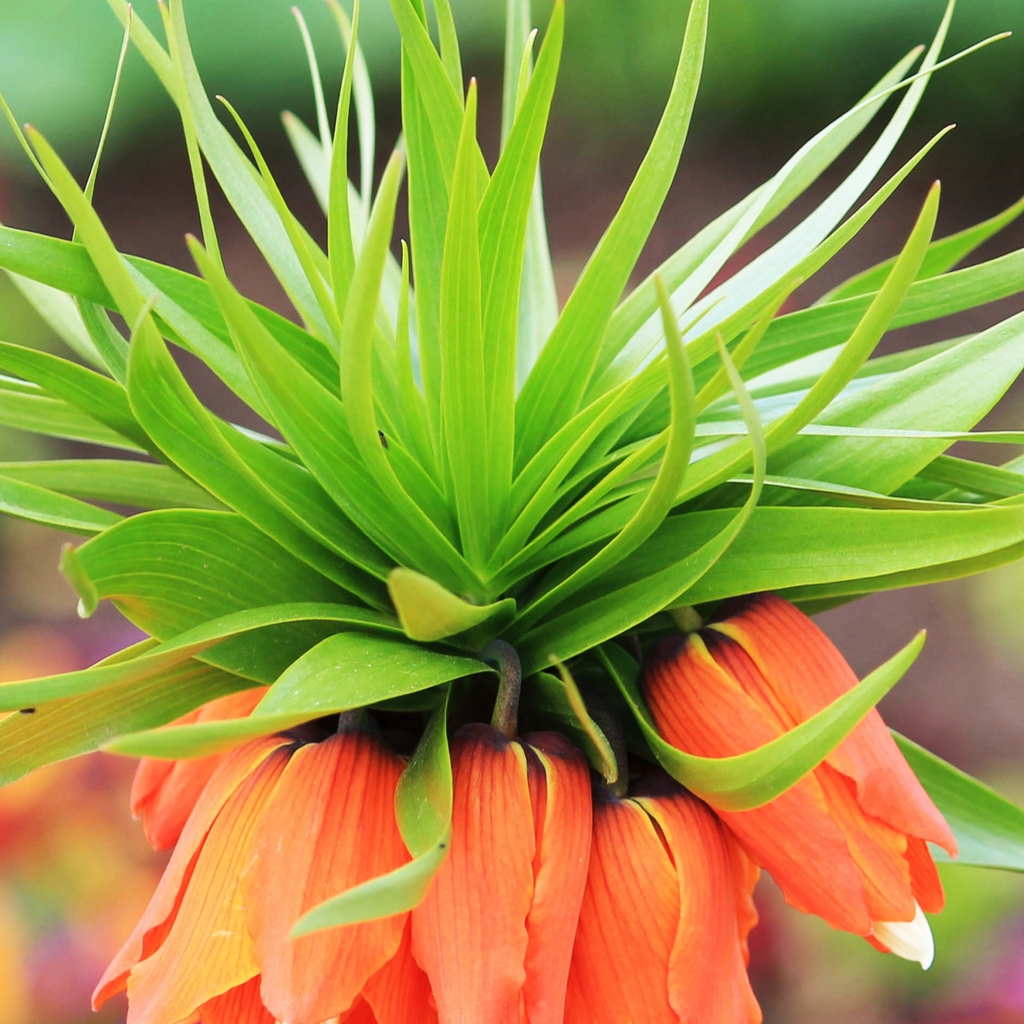}
    \caption{Prediction}
\end{subfigure} 
\rotatebox[origin=l]{90}{\makebox[0.13\textwidth][r]{Recovered}}

\end{center}
\caption{
Visual comparison of MSE heatmap images between the ground truth and predictions.}
\label{fig:hinet}
\end{figure*}

\subsection{Experiments on Image Hiding}
Table \ref{tab:hinet} compares the results of DLV-HiNet with 4bit-LSB, HiDDeN \cite{zhu2018hidden}, Baluja \cite{baluja2017hiding}, HiNet \cite{jing2021hinet} and HiNet$^\dagger$ quantitatively. For both cover/stego and secret/recovery image pairs of the three datasets, i.e., DIV2K, COCO and ImageNet, DLV-HiNet outperforms other methods in all four metrics.  Note that the numerical numbers of HiNet are worse than the ones in the original paper because we add quantization step in the training phase. The visual comparisons are shown in Fig. \ref{fig:hinet}. For better visual quality, we draw the heatmaps of the mean squared error between the ground truth and prediction. As can be seen, our DLV-HiNet can effectively reduce the errors in the edges and corners.

\section{Conclusion}
For bidirectional image rescaling models like IRN and HCFlow, a new latent variable is introduced to model the natural variations in image downscaling.
Combining with the original latent variable that models the variations of high-frequency details in image upscaling,
the dual latent variable enhancement is capable of further reducing expected restoration error in image upscaling.
With minimum impact on model complexity, this newly proposed enhancement is shown to able to improve restored HR image quality consistently for different test sets and testing scales while
maintaining high quality in downscaled LR images.  The DLV module is also shown to be effective in enhancing
image hiding models like HiNet, and potentially other INN-based restoration models.

\bibliographystyle{unsrt}  
\bibliography{mybib_2204}

\end{document}